\documentclass{article}

\usepackage{microtype}
\usepackage{graphicx}
\usepackage{subfigure}
\usepackage{booktabs} 

\usepackage{url}
\usepackage{amsmath,amssymb,mathtools,amsthm}
\usepackage{algorithm}
\usepackage{algpseudocode}
\usepackage{multirow}
\usepackage{colortbl}
\usepackage[export]{adjustbox}
\usepackage{tikz}
\usetikzlibrary{spy}
\usepackage{subcaption}
\usepackage{diagbox}
\usepackage{makecell}
\usepackage{enumitem}
\usepackage{listings}
\usepackage{xcolor}
\usepackage{svg}

\usepackage[accepted]{icml2025}

\usepackage[capitalize,noabbrev]{cleveref}

\theoremstyle{plain}

\theoremstyle{definition}

\theoremstyle{remark}

\usepackage[textsize=tiny]{todonotes}

\icmltitlerunning{TRISHUL}

\makeatletter
\renewcommand{\printAffiliationsAndNotice}[1]{%
    {\let\thefootnote\relax\footnotetext{%
        \\
        \textsuperscript{*}denotes equal contribution (alphabetical order). \\
        \textsuperscript{1} Fractal AI Research, India. \\
        \ifdefined\icmlcorrespondingauthor@text
          \textsuperscript{†} Correspondence to: \icmlcorrespondingauthor@text.
        \else
          {\bf AUTHORERR: Missing \textbackslash icmlcorrespondingauthor.}
        \fi
    }}%
}
\makeatother
\begin{document}

\twocolumn[
\icmltitle{TRISHUL: Towards Region Identification and Screen Hierarchy Understanding for Large VLM based GUI Agents}

\icmlsetsymbol{equal}{*}
\icmlsetsymbol{corr}{†}

\begin{icmlauthorlist}
\icmlauthor{Kunal Singh}{equal,comp,corr}
\icmlauthor{Shreyas Singh}{equal,comp}
\icmlauthor{Mukund Khanna}{comp}
\end{icmlauthorlist}

\icmlaffiliation{comp}{Fractal AI Research, Mumbai, India}

\icmlkeywords{Machine Learning, ICML}
\icmlcorrespondingauthor{Kunal Singh}{kunal.singh@fractal.ai}
\vskip 0.3in
]



\printAffiliationsAndNotice{} 

\begin{abstract}
Recent advancements in Large Vision Language Models (LVLMs) have led to the emergence of LVLM-based Graphical User Interface (GUI) agents developed under various paradigms. Training-based approaches, such as CogAgent and SeeClick, suffer from poor cross-dataset and cross-platform generalization due to their reliance on dataset-specific training. Generalist LVLMs, such as GPT-4V, utilize Set-of-Marks (SoM) for action grounding; however, obtaining SoM labels requires metadata like HTML source, which is not consistently available across platforms. Additionally, existing methods often specialize in singular GUI tasks rather than achieving comprehensive GUI understanding. To address these limitations, we introduce TRISHUL, a novel, training-free agentic framework that enhances generalist LVLMs for holistic GUI comprehension. Unlike prior works that focus on either action grounding (mapping instructions to GUI elements) or GUI referring (describing GUI elements given a location), TRISHUL seamlessly integrates both. At its core, TRISHUL employs Hierarchical Screen Parsing (HSP) and the Spatially Enhanced Element Description (SEED) module, which work synergistically to provide multi-granular, spatially, and semantically enriched representations of GUI elements. Our results demonstrate TRISHUL’s superior performance in action grounding across the ScreenSpot, VisualWebBench, AITW, and Mind2Web datasets. Additionally, for GUI referring, TRISHUL surpasses the ToL agent on the ScreenPR benchmark, setting a new standard for robust and adaptable GUI comprehension.
\end{abstract}

\section{Introduction}

\begin{figure*}
    \centering
    \includegraphics[width=1\linewidth]{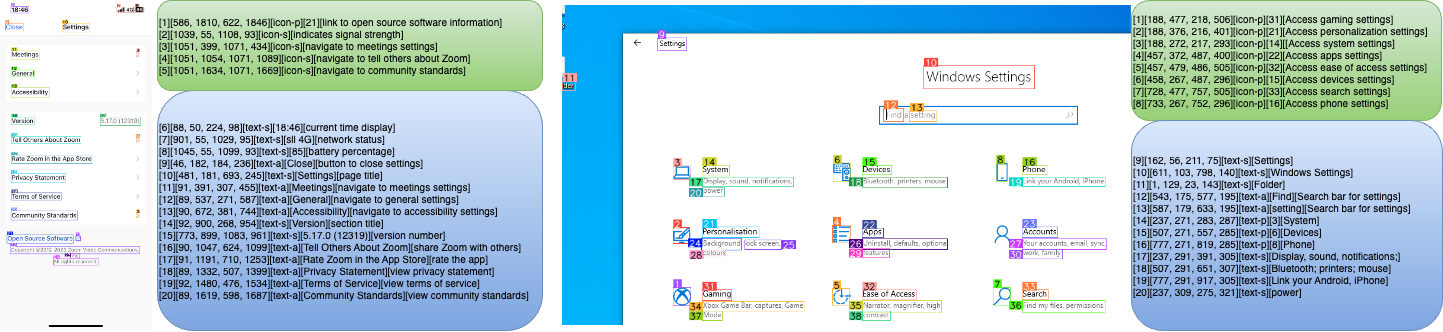}
    \caption{Screen parsing results showing detected GUI elements and their function descriptors leveraging our HSP and SEED modules}.
    \label{fig:prompt_groi}
\end{figure*}

\label{sec:intro}
Developing AI agents capable of operating digital devices through natural language commands has been a longstanding research goal \cite{pmlr-v70-shi17a, Liu2018ReinforcementLO, Gur2018LearningTN}. These agents can enhance productivity by automating tasks through Graphical User Interface (GUI). Early studies explored simplified settings \cite{pmlr-v70-shi17a, Liu2018ReinforcementLO, Gur2018LearningTN}, while later efforts \cite{Li2020MappingNL, Wang2021Screen2WordsAM, Li2020WidgetCG, He2020ActionBertLU, Bai2021UIBertLG, Wu2021ScreenPT, Zhang2021ScreenRC, Chen2020ObjectDF, Chen2020UnblindYA, Li2020MappingNL} leveraged GUI understanding to build more sophisticated agents. Recent approaches \cite{Yao2022WebShopTS, Gur2023ARW, Deng2023Mind2WebTA, zhou2023webarena, Sridhar2023HierarchicalPA} incorporate LLMs alongside structured GUI representations (e.g., HTML, DOM trees, View Hierarchy) to enhance comprehension.

With advances in LVLMs, studies \cite{zheng2023seeact, Deng2023Mind2WebTA, He2024WebVoyagerBA, Zhang2023AppAgentMA, Furuta2023MultimodalWN} have integrated visual perception to improve performance on benchmarks like Mind2Web \cite{Deng2023Mind2WebTA} and WebArena \cite{zhou2023webarena}. However, these models struggle with visual grounding \cite{yang2023setofmark}, relying heavily on structured metadata, which is often unavailable, noisy, or misaligned. SeeAct \cite{zheng2023seeact} improves action grounding in GPT-4V \cite{GPT-4V} via set-of-marks (SoM) \cite{yang2023setofmark}, but its dependency on structured data introduces limitations.


\subsection{Related Works \& Motivation}

Recent research has focused on developing agents that rely solely on visual perception to interact with GUIs in a human-like manner. These works on purely vision-based GUI agents using LVLMs have evolved along 2 main approaches:

\textbf{End to End Training based GUI Agents}: Multiple studies \cite{Hong2023CogAgentAV, You2024FerretUIGM, Cheng2024SeeClickHG, Bai2024DigiRLTI, shaw2023pixels} have trained LVLMs on GUI navigation tasks for various platforms/device-types. 



\textbf{Test-time assistance with visual perception tools}: Studies have leveraged visual perceptions tools to assist generalist LVLMs like GPT-4V. MM-Navigator \citep{Yan2023GPT4VIW} leverages pre-trained icon detector module. A concurrent work to ours, Omniparser \citep{OmniParser}, trains a YOLO-v8 \citep{yolov8_ultralytics} based icon detection \& BLIPv2 \citep{blip2} based icon captioner modules for action grounding. Tree-of-Lens (ToL) Agent \cite{fan2024readpointedlayoutawaregui} trains a perception module for GUI referring task of generating region description based on user selected point. 

Multiple GUI navigation-related benchmarks \cite{Liu2024VisualWebBenchHF, Xie2024OSWorldBM} and studies \cite{zheng2023seeact, Cheng2024SeeClickHG} have highlighted two major weaknesses among pure vision-based GUI navigation agents. Firstly, the performance of these methods trained on certain distribution of user interfaces don't generalize well across platforms/device types. Given the rapid pace with which new user interfaces are introduced every day, the generalizability of training based approaches to Out-Of-Distribution samples remains a challenge. Secondly, most of the GUI agents such as DigiRL \cite{Bai2024DigiRLTI}, SeeClick \cite{Cheng2024SeeClickHG},  MM-Navigator \cite{Yan2023GPT4VIW} are optimized for specialized GUI related tasks (majorly action prediction \& grounding), and often evaluate on diversely sourced but thematically similar tasks and metrics, hence they lack proper GUI comprehension capabilities across different tasks and interfaces.

\begin{algorithm}
\caption{Hierarchical Screen Parsing}\label{alg:groi}
\scriptsize  
\begin{algorithmic}[1]
    \Require Image $I$, $A_{\textit{thresh-GROI}}$, $A_{\textit{thresh-Icon}}$, $IOU_{\textit{thresh}}$, \textit{SAM}, \textit{OCR}
   
    \State Initialize: \textit{SAM}, \textit{OCR}, $A_{\textit{thresh}}$, $IOU_{\textit{thresh}}$
    \State Sample N points $\mathcal{P} \gets \mathcal{U}(0, W) \times \mathcal{U}(0, H)$ \Comment{Image Size ($W$, $H$)}
    \State $\mathcal{B} \gets \textit{SAM}(I, \mathcal{P}), \quad \mathcal{T} \gets \textit{OCR}(I)$ \Comment{SAM boxes $\mathcal{B}$ and OCR boxes $\mathcal{T}$}
    
    \State Initialize $\mathcal{G} \gets \emptyset, \mathcal{I} \gets \emptyset$ \Comment{GROI candidates and Icon candidates}
    
    \For{each $b \in \mathcal{B}$}
        \If{Area$(b) > A_{\textit{thresh-GROI}}$}
            \State $\mathcal{G} \gets \mathcal{G} \cup \{b\}$  \Comment{Add to GROI candidates}
        \EndIf
        \If{Area$(b) < A_{\textit{thresh-Icon}}$}
            \State $\mathcal{I} \gets \mathcal{I} \cup \{b\}$ \Comment{Add to Icon candidates}
        \EndIf
    \EndFor
    
    \State Initialize $\mathcal{S} \gets \emptyset$ \Comment{Information Scores for Non Max Suppression (NMS)}
    \State $\mathcal{I}_{\text{filtered}}, \mathcal{T}_{\text{filtered}} \gets$ \textit{Overlap Removal and Filtering}($\mathcal{I}, \mathcal{T}$)
    
    \For{each $b \in \mathcal{G}$}
        \State $\mathcal{N}_{\text{inside}} = |\{ \mathcal{T}_b^\text{inside} \}| + |\{ \mathcal{I}_b^\text{inside} \}|$ \Comment{Number of boxes  inside $b$}
        
        \State $\mathcal{N}_{\text{inter}}  = |\{ \mathcal{T}_b^\text{intersect} \}| + |\{ \mathcal{I}_b^\text{intersect} \}|$
        \Comment{Number of boxes intersecting $b$}
        
        \State $\mathcal{S} \gets \mathcal{S} \cup \left\{ 
        \frac{\mathcal{N}_{\text{inside}}}
        {\sqrt{1 + \mathcal{N}_{\text{inter}} \cdot \text{Area}(b)}} 
        \right\}$ \Comment{Information Score for $b$}    
    \EndFor

    \State $\mathcal{G}_{\text{filtered}} \gets$ \textit{NMS}($\mathcal{G}, \mathcal{S}, IOU_{\textit{thresh}}$) 
    \Comment{Apply NMS to get Filtered GROIs}

    \State \textbf{return} $\mathcal{G}_{\text{filtered}}, \mathcal{I}_{\text{filtered}}, \mathcal{T}_{\text{filtered}}$
\end{algorithmic}
\end{algorithm}

\begin{figure*}
    \centering
    \includegraphics[width=0.8\linewidth]{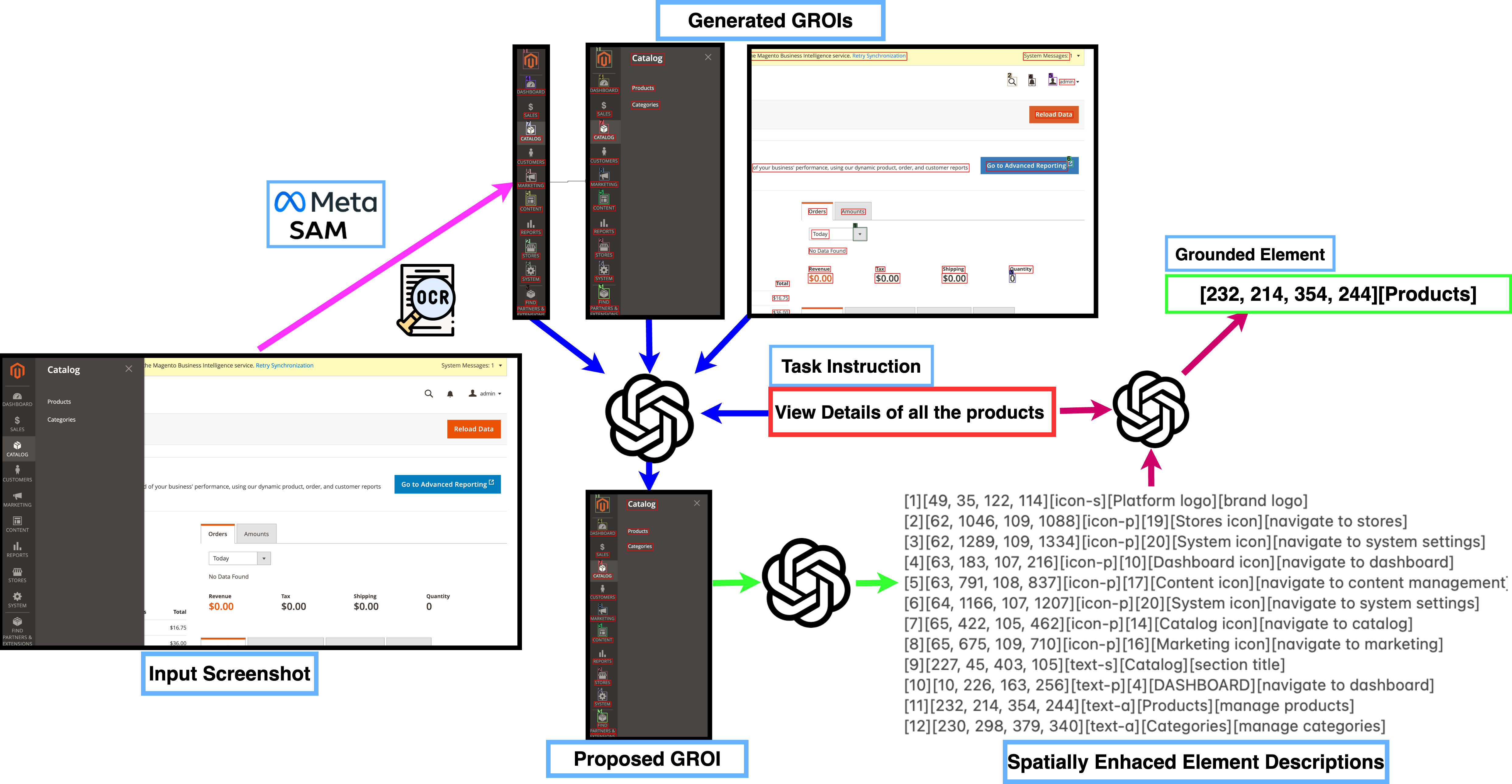}
    \caption{TRISHUL: Agentic Action Grounding Framework, Pink arrow, denotes our Hierarchical Screen Parsing (HSP) method, to generate GROIs and local element annotations, Green arrows represent our Spatially Enhanced Element Descriptor (SEED) workflow, Blue arrows represent our GROI proposal framework and Magenta Arrow shows, the Set of Marks (SoM) based  Grounding workflow.}
    \label{fig:main_fig}
\end{figure*}

\subsection{Contribution}

To address these challenges, we introduce TRISHUL, a training-free, agentic framework for comprehensive GUI screen understanding. TRISHUL equips LVLMs with the capabilities required to perform diverse GUI interaction tasks, it utilizes foundational models to parse and build a rich hierarchical understanding of the GUI screens,to enhance their action grounding and GUI referring capabilities.

\textbf{Hierarchical Screen Parsing (HSP):} The HSP module organizes GUI elements across two distinct levels of granularity: broad regions called Global Regions of Interest (GROIs) which cluster related components and local elements like icons, text, and images. This hierarchical structuring captures spatial and semantic relationships between different GUI components, providing a multi-layered comprehensive GUI screen understanding.

\textbf{Spatially Enhanced Element Description (SEED):} SEED generates contextually aware and spatially informed functionality descriptions for local elements by analyzing their relative positioning with respect to other elements in the GUI. By associating nearby icons and text, SEED enables the generation of high-fidelity functionality descriptions for GUI elements, facilitating a more nuanced understanding of each element's role.

We evaluate TRISHUL on ScreenSpot \cite{Cheng2024SeeClickHG}, VisualWebBench \cite{Liu2024VisualWebBenchHF}, Mind2Web \cite{Deng2023Mind2WebTA}, and AITW \cite{rawles2023androidwildlargescaledataset}, demonstrating that GPT-4V \cite{GPT-4V} and GPT-4o \cite{gpt4o} using TRISHUL surpass prior state-of-the-art methods in action grounding and episodic instruction-following tasks. Additionally, we validate TRISHUL’s effectiveness in GUI referring via the Screen PR dataset, improving accessibility applications and user interaction feedback





\section{Methodology}

This section outlines the design of our training-free screen comprehension modules, HSP and SEED, and sheds light on their integration into our action grounding and GUI referring agent.

\subsection{Hierarchical Screen Parsing}
\label{sec:HSP}

The hierarchical screen parsing process is formalized in Algorithm \ref{alg:groi}. Initially, the screen image \( I \) is passed through SAM \cite{SAM} and EasyOCR \cite{EasyOCR}. The generated bounding boxes are filtered based on predefined area thresholds \( A_{\text{thresh-GROI}} \) and \( A_{\text{thresh-LE}} \) to generate  GROI candidates and Local Elements (LE). Local Elements collectively refer to bounding boxes for text, icon, buttons and images in the GUI. We then apply an overlap removal and filtering function to refine the icon and text bounding boxes by removing redundant and unwanted local elements. 

For each GROI candidate, the number of boxes inside and intersecting with the GROI is calculated. An Information Score \( \mathcal{S} \) is then computed for each candidate based on the ratio of the number of bounding boxes inside, to the area of the GROI, adjusted by the number of intersecting boxes. This score provides a measure of the GROI's information content, helping the system to prioritize larger and more informative regions for inclusion in the hierarchical tree.

Finally, a Non-Max-Suppression (NMS) algorithm is applied to the GROI candidates based on their Information Scores. The resulting filtered set of GROIs, icons, and text boxes are returned as the final hierarchical structure, which contains all the relevant GUI elements grouped together through GROIs. For specific details on the Overlap Removal, Filtering and NMS algorithm refer to Appendix \ref{app: HSP_details}

\begin{figure*}
    \centering
    \includegraphics[width=0.8\linewidth]{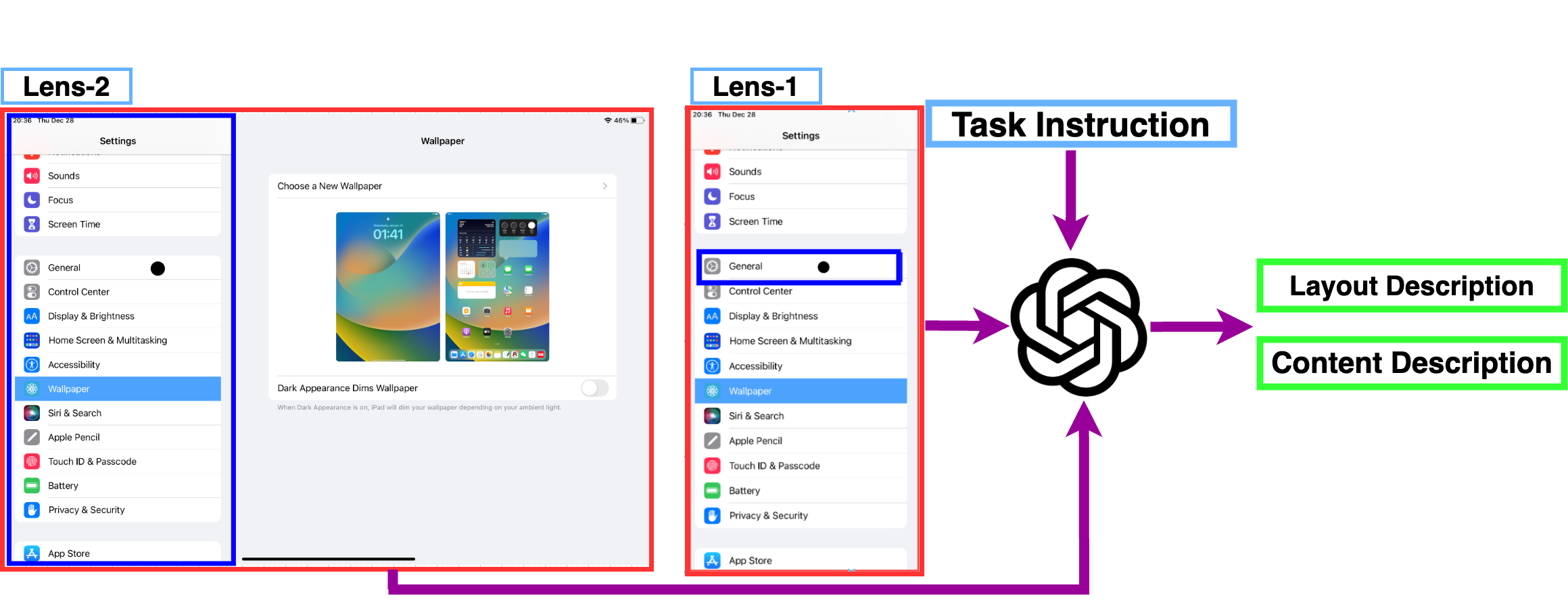}
    \caption{TRISHUL: Agentic GUI Referring Framework, the 2 Lenses created using our HSP module for local and global context. Lens-1 contains the local element (blue) in the cropped GROI (red), Lens-2 contains the GROI (blue) in the full input screenshot (red).The selected point is represented as the black dot. Both lenses are fed to the LVLM to generate Layout and Task description.}
    \label{fig:ungabunga}
\end{figure*}

\subsection{SEED: Spatially Enhanced Element Description Generation}
\label{sec:SEED}

Accurately describing the functionality of local GUI elements is essential for effective understanding of GUI and action grounding. Relying solely on visual appearance is unreliable since identical icons can serve different purposes in different contexts, and distinct icons may represent similar functions, leading to ambiguity.  Textual and semantic cues around GUI elements help clarify functionality. Pairing icons with nearby text enables precise descriptions, while semantic associations (e.g., text linked to input fields or buttons) aid in identifying actionable elements.  

We introduce SEED (Spatially Enhanced Element Description), a prompting framework that employs Chain of Thought (CoT) \cite{wei2023chainofthoughtpromptingelicitsreasoning} and In-Context Learning (ICL) \cite{Brown2020LanguageMA} to generate spatially and semantically informed functional descriptions for all GUI elements. SEED processes an image \( I \) annotated with SoM-style ID tags, and a prompt with bounding boxes for detected elements (via our HSP module), and OCR-extracted text descriptors:  

\begin{equation}
    \mathcal{B}_\text{icon} = \{ (i, b_{\text{icon},i}) \}_{i=1}^{N_\text{icon}} 
\end{equation}
\begin{equation}
    \mathcal{B}_\text{text} = \{ (i, b_{\text{text},i}, d_i) \}_{i=N_\text{icon}}^{N_\text{total}},
\end{equation}  

where \( b_{\text{icon},i} \) and \( b_{\text{text},i} \) are bounding boxes, and \( d_i \) represents OCR-derived text descriptors.  

SEED outputs a spatially enhanced descriptor set \( \mathcal{A} \):  

\begin{equation}
\mathcal{A} = \left\{ b, \ell, a, d \mid b \in \mathcal{B}_\text{icon} \cup \mathcal{B}_\text{text} \right\}
\end{equation}  

Each element's attributes include bounding box \( b \), label \( \ell \in \{ paired, standalone, picture, actionable-text \} \), set of associated elements \( a \), and a spatially enhanced functional description \( d \).  

SEED classifies elements as paired or standalone based on semantics and positioning. Paired elements combine descriptors from nearby text/icons for a unified description, while standalone elements rely on visual cues alone. Text elements linked to interactive components (e.g., input fields, search bars, buttons) are labeled as actionable, and embedded icons are classified as \{picture\}.  

We use ICL \cite{Brown2020LanguageMA} with six examples from the ScreenSpot \cite{Jurmu2008ScreenSpotMR} dataset,  The full SEED prompt with specific details about the SEED module is available  in Appendix \ref{sec:appendix}.

\subsection{Agentic Formulation of Action Grounding}
\begin{table}[h!]
\label{GROI_prop_acc}
    \centering
    \begin{tabular}{lcccc}
        \toprule
        \multirow{2}{*}{Platform} & \multicolumn{2}{c}{ScreenSpot} & \multicolumn{2}{c}{VisualWebBench} \\
        \cmidrule(lr){2-3} \cmidrule(lr){4-5}
        & GPT-4o & GPT-4V & GPT-4o & GPT-4V \\
        \midrule
        Mobile & 0.91 & 0.81 & - & - \\
        Web & 0.96 & 0.83 & 0.93 & 0.86 \\
        PC & 0.92 & 0.83 & - & - \\
        Overall & 0.93 & 0.82 & 0.93 & 0.86 \\
        \bottomrule
    \end{tabular}
    \caption{GROI proposal accuracy.}
    \label{groi_prop}
\end{table}

\label{sec:agent_framework}

This section explains how the hierarchical nature of GUIs is leveraged for enhanced SoM style action grounding in LVLMs as explained in fig. \ref{fig:main_fig}. Given an image \( I \) with Global Regions of Interest (GROIs) \( \mathcal{G} \), bounding boxes for icons \( \mathcal{B}_{\text{icon}} \) and text \( \mathcal{B}_{\text{text}} \), OCR-derived text descriptors \( d_j \), and an instruction \( I_s \), the task is to identify the bounding box \( \mathcal{B} \) corresponding to the correct element required to complete the instruction in a single step. 

TRISHUL performs action grounding in two stages. First, it proposes the most relevant GROI by passing the full annotated image \( I_{\text{annotated}} \), cropped GROIs \( \mathcal{G}_{\text{cropped}} \), and instruction \( I_s \) to the LVLM. The model outputs descriptions \( \mathcal{D}_\mathcal{G} \) for each GROI and the ID of the most relevant one:  

\begin{equation}
    \{ I_{\text{annotated}}, \mathcal{G}_{\text{cropped}}, I_s \} \longrightarrow \left\{ \mathcal{D}_\mathcal{G}, \, \text{ID}_{\text{GROI}} \right\}.
\end{equation}

GROI proposal accuracy is evaluated by checking if the ground truth bounding box midpoint lies inside the proposed GROI. Results with GPT-4o and GPT-4V on ScreenSpot \cite{Jurmu2008ScreenSpotMR} and VisualWebBench \cite{Liu2024VisualWebBenchHF} (Table \ref{groi_prop}) confirm the effectiveness of our GROI ranking module.  

Next, we use SEED (Section \ref{sec:SEED}) to generate functionality descriptors for all local elements in the proposed GROI. The annotated image and descriptors are then used in a Set of Marks \cite{yang2023setofmark} framework to predict the bounding box for grounding the instruction.  

\subsection{Agentic Formulation of GUI referring task}

In this section we describe how the hierarchical screen parsing module can be leveraged to increase the ability of LVLMs on the GUI referring task as explained in fig. \ref{fig:ungabunga}. Given the input GUI screenshot \textit{I}, the task involves describing the content and layout of any point \( P_{\text{i}} \) on the screen as input by a user, we use the input screenshot to detect all local elements and corresponding GROI candidates. We then identify the bounding box of the local element containing the selected point, and then the GROI encompassing this local element. Following the prompting approach of the ToL agent in \cite{fan2024readpointedlayoutawaregui},  we curate two “lenses” or images to illustrate this hierarchy. The first lens consists of only the GROI region cropped from the original image, highlighting the local element with a labeled bounding box and marking the input point. The second lens shows the complete screenshot, highlighting the GROI with a labeled bounding box. Both lenses, along with the point coordinate \( P_{\text{i}} \) and input prompt, are sent to an LVLM, to generate the content description \( \hat{D}_{\text{c}} \) and the layout description \( \hat{D}_{\text{l}} \). 

\section{Experiments}

\subsection{ScreenSpot and VisualWebBench}
\label{subsec:screenspot_vwb}

\begin{table*}[ht]
\small
  \centering
  \begin{tabular}{@{}lcccccccc@{}}
    \toprule
    \textbf{Method} & \multicolumn{2}{c}{\textbf{Mobile (ScreenSpot)}} & \multicolumn{2}{c}{\textbf{Desktop (ScreenSpot)}} & \multicolumn{2}{c}{\textbf{Web (ScreenSpot)}} & \textbf{ScreenSpot} & \textbf{VisualWebbench} \\
    \cmidrule(lr){2-3} \cmidrule(lr){4-5} \cmidrule(lr){6-7}
     & \textbf{Text} & \textbf{Icon/widget} & \textbf{Text} & \textbf{Icon/widget} & \textbf{Text} & \textbf{Icon/widget} & \textbf{Overall} & \textbf{Overall} \\
    
    \midrule
    \multicolumn{9}{l}{\textbf{Training Based}} \\
    SeeClick & 78.0 & 52.2 & 72.2 & 30.0 & 55.7 & 32.5 & 53.4 & 31.0 \\
    CogAgent & 67.0 & 24.0 & 74.2 & 20.0 & 70.4 & 28.6 & 47.4 & 59.0 \\
    OmniParser (GPT-4V) & 90.1 & 54.1 & 88.6 &  60.0 & 73.4 & 27.1 & 66.9 & 58.3 \\
    OmniParser$^*$(GPT-4V)  & 92.1 & 55.2 & 90.1 &  61.1 & 77.4 & 30.1 & 69.5 & 63.1 \\
    OmniParser (GPT-4o) & \textbf{93.9} & \textbf{57.0} & \textbf{91.3} & \textbf{63.6} & \textbf{\textit{81.3}} & \textbf{51.0} & \textbf{72.6} & \textbf{68.9} \\
    OmniParser$^*$(GPT-4o)& \textbf{\textit{94.8}} & \textbf{\textit{66.3}} & \textbf{\textit{95.4}} & \textbf{\textit{64.2}} & \textbf{80.8} & \textbf{32.0} & \textbf{\textit{73.7}} & \textbf{\textit{69.9}} \\

    \midrule
    \multicolumn{9}{l}{\textbf{Training Free}} \\
   
    GPT-4V & 22.6 & 24.5 & 20.2 & 11.8 & 9.2 & 8.8 & 16.2 & 6.0 \\
    GPT-4o & 20.2 & 24.9 & 21.1 & 23.6 & 12.2 & 7.8 & 18.2 & 6.7 \\
   
    TRISHUL$^\dagger$ (GPT-4V) & 75.8 & 38.4 & 66.3 & 25.4 & 69.5 & 31.2 & 53.4 & 56.3 \\
    TRISHUL$^*$ (GPT-4V) & 88.6 & 37.9 & 82.9 & 23.5 & 72.6 & 29.1 & 59.0 & 58.1 \\
    TRISHUL$^{*\dagger}$ (GPT-4V) & 86.0 & {43.7} & 77.3 & {32.8} & {75.2} & {40.8} & {61.9} & \textbf{{68.0}} \\
   
    TRISHUL$^\dagger$ (GPT-4o) & 92.1 & \textbf{63.4} & 83.7 & 38.2 & 80.2 & \textbf{42.1} & 69.3 & 60.2 \\
    TRISHUL$^*$ (GPT-4o) & \textbf{92.7} & 62.0 & \textbf{\textit{90.2}} & \textbf{39.2} & \textbf{\textit{84.8}} & 40.8 & \textbf{71.1} & 62.1 \\
    TRISHUL$^{*\dagger}$ (GPT-4o) & \textbf{\textit{93.8}} & \textbf{\textit{64.6}} & \textbf{85.6} & \textbf{\textit{45.7}} & \textbf{83.5} & \textbf{\textit{44.7}} & \textbf{\textit{72.2}} & \textbf{\textit{68.0}} \\
   
    \bottomrule
  \end{tabular}
  \caption{Performance across platforms and methods on ScreenSpot (Mobile, Desktop, Web) and VisualWebbench datasets. 
  $^*$ denotes the usage of SEED module to improve the element functionality descriptors generated using OCR (for TRISHUL) / BLIPv2 (for OmniParser). 
  $^\dagger$ represents GROI-based action grounding instead of using the full image. 
  $^{*\dagger}$ represents our proposed end-to-end framework for action grounding that uses GROIs and SEED descriptors. 
  Refer to Sec. \ref{subsec:screenspot_vwb} for detailed discussion.}
  \label{tab:screenspot}
\end{table*}

\textbf{Dataset and Experiments}- We evaluate the action grounding capability of TRISHUL agent on the ScreenSpot \cite{Jurmu2008ScreenSpotMR} dataset. ScreenSpot consists of 610 interface screenshots from mobile (iOS, Android), desktop (macOS, Windows), and web platforms, paired with 1,276 task instructions corresponding to actionable GUI elements. Traditional training-based methods, which are often trained on datasets like Screenspot, tend to perform poorly on out-of-distribution samples such as those from VisualWebBench due to domain shift. Therefore, to assess the generalization capability of our approach, we also utilize the VisualWebBench \cite{Liu2024VisualWebBenchHF} dataset's action grounding subset, which consists of 103 pairs of images and their corresponding instruction. 

\textbf{Implementation Details:}  
The formulation of the action grounding tasks for the datasets used in our experiments is discussed in detail in Section \ref{sec:agent_framework}. The specific prompts employed for these tasks are provided in the Appendix (Figure \ref{fig:prompt_screenspot}).

Unfortunately, we were unable to replicate the results reported by OmniParser in their study on the ScreenSpot benchmark using the publicly available weights and codebase. In Table \ref{tab:screenspot}, we present the performance metrics for OmniParser as obtained from our own experiments on the ScreenSpot and VisualWebBench datasets. Due to the non-reproducibility of their results as observed above and limited resources, we were unable to verify their results on the AiTW and Mind2Web benchmarks hence we have chosen to exclude their results for these benchmarks from our analysis.

\textbf{Evaluation and Results}: As shown in Table \ref{tab:screenspot}, the TRISHUL agent, when paired with LVLMs (GPT-4V \cite{GPT-4V} and GPT-4o \cite{gpt4o}), significantly outperforms the baseline GPT-4V and GPT-4o. Our approach also surpasses task-specific models such as SeeClick \cite{Cheng2024SeeClickHG} and CogAgent \cite{Hong2023CogAgentAV}, achieving an overall accuracy of 61.9\% with GPT-4V and 72.2\% with GPT-4o on the ScreenSpot benchmark. This performance exceeds SeeClick’s 53.4\%, CogAgents 47.4\% and closely rivals OmniParser’s 72.6\%. On VisualWebBench \cite{Liu2024VisualWebBenchHF}, unlike SeeClick, which suffers a sharp drop in accuracy on out-of-distribution data with 31\% accuracy, TRISHUL maintains strong generalization, achieving a robust 68.0\% accuracy with both GPT-4V and GPT-4o closely matching the performance of OmniParser which achieves 68.9\%.

We further present ablations in Table \ref{tab:screenspot} to assess the impact of the SEED module and GROI-based action grounding in TRISHUL. Removing SEED (TRISHUL$^\dagger$) results in a notable accuracy drop of 8.5\% for GPT-4V and 2.9\% for GPT-4o on ScreenSpot. Similarly, eliminating GROI-based action grounding (TRISHUL$^*$) reduces accuracy by 2.9\% for GPT-4V and 1.1\% for GPT-4o. These results highlight the critical role of these components in TRISHUL’s performance.  

Additionally, we demonstrate TRISHUL’s modularity by integrating its components into existing grounding pipelines. In Table \ref{tab:screenspot}, we show that augmenting OmniParser’s BLIPv2-derived icon descriptors—originally lacking local semantic context—with TRISHUL’s SEED module (OmniParser$^*$) yields the best performance among training-based methods.  

Our GROI-based action grounding proves particularly effective for web and desktop platforms, where hierarchical and content-dense GUIs benefit from structured decomposition. However, its impact is less pronounced in mobile interfaces, where regions have minimal semantic separation. Further details can be found in Appendix \ref{sec6.3}.  Lastly, we observe that GPT-4o outperforms GPT-4V significantly when paired with SEED, suggesting that improved reasoning capabilities in LVLMs enhance the accuracy of SEED-generated descriptions.

\begin{table*}[ht]
  \centering
  
  \begin{tabular}{@{}lccccccc@{}}
    \toprule
    \textbf{Method} & \textbf{General} & \textbf{Install} & \textbf{GoogleApps} & \textbf{Single} & \textbf{WebShopping} & \textbf{Overall} \\
    \midrule
    ChatGPT-CoT & 5.9 & 4.4 & 10.5 & 9.4 & 8.4 & 7.7 \\
    Palm2-CoT & - & - & - & - & - & 39.6 \\    GPT-4V + Image & 41.7 & 42.6 & 49.8 & 72.8 & 45.7 & 50.5 \\
    MM-Navigator (GPT-4V) & 43 & 49.2 & 46.1 & \textbf{\textit{78.3}} & 48.2 & 53.0 \\
    MM-Navigator (GPT-4o) & \textbf{\textit{55.8}} & 58.2 & 48.2 & 76.9 & 52.1 & 57.8 \\
    SeeClick (Qwen-VL) & \textbf{54.0} & \textbf{\textit{66.4}} & \textbf{54.9} & 63.5 & \textbf{\textit{57.6}} & \textbf{59.3}\\
    \midrule
    TRISHUL (GPT-4V) & 47.5 & 50.7 & 50.7 & 66.7 & 49.5 & 54.5 \\
    TRISHUL (GPT-4o) & 52.9 & \textbf{60.7} & \textbf{\textit{55.0}} & \textbf{78.2} & \textbf{52.6} & \textbf{\textit{60.0}} \\
    \bottomrule
  \end{tabular}
  \caption{Results on the different categories on the AITW dataset. TRISHUL (GPT-4V) outperforms all prior GPT-4V  baselines that use IconNet's  element detections. TRISHUL (GPT-4o) outperforms TRISHUL (GPT-4V) by 5.55\% achieving State of the Art performance.}
  \label{tab:AITW}
\end{table*}

\begin{table*}[ht]
\centering
\small
\begin{tabular}{lcccccccccccc}
\toprule
\textbf{Methods} & \textbf{Modality} & \multicolumn{3}{c}{\textbf{Cross-Website}} & \multicolumn{3}{c}{\textbf{Cross-Domain}} & \multicolumn{3}{c}{\textbf{Cross-Task}} \\
\cmidrule(lr){3-5} \cmidrule(lr){6-8} \cmidrule(lr){9-11}
& & \textbf{Ele.Acc} & \textbf{Op.F1} & \textbf{Step SR} & \textbf{Ele.Acc} & \textbf{Op.F1} & \textbf{Step SR} & \textbf{Ele.Acc} & \textbf{Op.F1} & \textbf{Step SR} \\
\midrule
MindAct (gen) & HTML & 13.9 & 44.7 & 11.0 & 14.2 & 44.7 & 11.9 & 14.2 & 44.7 & 11.9 \\
MindAct & HTML & 42.0 & 65.2 & 38.9 & 42.1 & 66.5 & 39.6 & 42.1 & 66.5 & 39.6 \\
GPT-3.5-Turbo & HTML & 19.3 & 48.8 & 16.2 & 21.6 & 52.8 & 18.6 & 21.6 & 52.8 & 18.6 \\
GPT-4 & HTML & 35.8 & 51.1 & 30.1 & 37.1 & 46.5 & 26.4 & 41.6 & 60.6 & 36.2 \\
GPT-4V+Text & HTML, Image & 38.0 & 67.8 & 32.4 & 42.4 & 69.3 & 36.8 & 46.4 & 73.4 & 40.2 \\
\midrule
GPT-4V+SOM & Image & - & - & 32.7 & - & - & 23.7 & - & - & 20.3 \\
CogAgent & Image &18.4 & 42.2 & 13.4 & 20.6 & 42.0 & 15.5 & 22.4 & 53.0 & 17.6 \\
Qwen-VL & Image & 13.2 & \textbf{83.5} & 9.2 & 14.1 & 84.3 & 12.0 & 14.1 & 84.3 & 12.0 \\
SeeClick & Image & 21.4 & 80.6 & 16.4 & 23.2 & \textbf{84.8} & 20.8 & 28.3 & \textbf{87.0} & 25.5 \\
\midrule
TRISHUL (GPT-4V) & Image & 33.91 & 74.33 & 27.98 & 36.49 & 76.60 & 31.71 & 34.04 & 71.88 & 29.76 \\ 
TRISHUL (GPT-4o) & Image & \textbf{31.43} & 81.52 & \textbf{24.53} & \textbf{37.12} & 82.96 & \textbf{32} & \textbf{37.58} & 83.78 & \textbf{32.52} \\ 
\bottomrule
\end{tabular}

\caption{Results for Cross-Website, Cross-Domain, and Cross-Task scenarios with Element Accuracy, Operational F1, and Step Success Rate metrics on the Mind2Web benchmark. TRISHUL (GPT-4o) consistently gives better Element Accuracy and Step Success Rate in all three scenarios on Image modality, its performance
trails state-of-the-art HTML-based method like MindAct}
\label{tab:mind2web}
\end{table*}

\begin{table}[ht]
  \centering
  \small 
  \setlength{\tabcolsep}{2pt} 
  \begin{tabular}{@{}llcccc@{}}
    \toprule
    \textbf{LVLM} & \textbf{Method} & \textbf{Desc. Acc.} & \textbf{Cont. Acc.} & \textbf{BERT} & \textbf{ROUGE} \\
    \midrule
    \multirow{3}{*}{GPT-4V} & Baseline & 8 & 0.92 & 0.7130 & 0.1462 \\
                            & ToL & 31.84 & 14.24 & \textbf{0.7230} & 0.1527 \\
                            & \textbf{TRISHUL} & \textbf{32.64} & \textbf{17.07} & 0.7220 & \textbf{0.1534} \\
    \midrule
    \multirow{3}{*}{Claude-3.5} & Baseline & 16.04 & 7.43 & 0.7274 & 0.1134 \\
                            & ToL & 60.56 & 43.02 & 0.7306 & 0.1462 \\
                            & \textbf{TRISHUL} & \textbf{60.91} & \textbf{49.74} & \textbf{0.7336} & \textbf{0.1495} \\
    \midrule
    \multirow{3}{*}{GPT-4o} & Baseline & 18.82 & 5.64 & 0.6948 & 0.1843 \\
                            & ToL & 71.30 & 42.46 & 0.7147 & 0.1869 \\
                            & \textbf{TRISHUL} & \textbf{71.58} & \textbf{43.59} & \textbf{0.7151} & \textbf{0.1871} \\
    \bottomrule
  \end{tabular}
  \caption{Evaluation of description and content accuracy, BERT score, and ROUGE-L score across different methods on the Screen Point-and-Read benchmark. Desc. Acc. - Description Accuracy, Cont. Acc. - Content Accuracy}
  \label{tab:screenpr}
\end{table}

\subsection{AITW}

\textbf{Dataset and Experiments} To evaluate TRISHUL on the mobile navigation benchmark AITW\cite{rawles2023androidwildlargescaledataset}, which consists of 30,000 instructions and 715,000 trajectories, we use the same train/test split as defined in \cite{Cheng2024SeeClickHG}. This split retains only one trajectory per instruction, ensuring no overlap between the train and test sets.

\textbf{Implementation details}- We adopt a similar prompt format to that used in MM-Navigator \cite{Yan2023GPT4VIW}, where we label the detected elements on the screen using SoM prompting and present the model with the annotated image and the clean image. However, we replace IconDet's bounding boxes (as used in MM-Navigator)  with local element boxes generated from our Hierarchal Screen Parsing method, and also provide our spatially enhanced element descriptions (Section \ref{sec:SEED}) for all the local elements in our input prompt. The exact prompt is mentioned in the Appendix in Figure \ref{fig:prompt_aitw}

\textbf{Evaluation and Results} In Table \ref{tab:AITW}, we report the baselines as presented in MM-Navigator\cite{Yan2023GPT4VIW}. The best performing baseline incorporates action history and uses only image modality for navigation. MM-Navigator presents baselines with GPT-4V only, we also run MM-navigator's best configuration (Image+History) with GPT-4o to contrast it with TRISHUL's GPT-4o performance.
We observe that TRISHUL with GPT-4V outperforms all prior GPT-4V-based baselines, achieving an overall accuracy of 54.5\%. With GPT-4o model, TRISHUL achieves an average accuracy of 60\%, surpassing MM-Navigator's GPT-4o baseline by over 2.2\% to become the state of the art. 

\subsection{Mind2Web}
\textbf{Dataset and Experiments}- 
To test on the web-navigation task we use the Mind2Web \cite{Deng2023Mind2WebTA} dataset. The test set consists of three different categories - Cross Task, Cross Website, and Cross Domain having 252, 177, and 912 tasks respectively.

\textbf{Implementation details} -  We use the pre-processed test set provided by \cite{Yan2023GPT4VIW}. During inference, we feed the detected local elements outputs from our Hierarchical Screen Parsing (HSP) module along with the clean image. Additionally, our input prompts are augmented with the descriptions of local elements from our SEED module. The prompt is mentioned in the Appendix in Figure \ref{fig:prompt_mind2web}

\textbf{Evaluation and Results} - The results are presented in Table \ref{tab:mind2web} where we compare multiple baselines across two modalities HTML and image. GPT-4V+SoM and GPT-4V+Text correspond to SeeAct \cite{zheng2023seeact} with image annotations and text choice grounding methods respectively. Without using any parsed HTML information, TRISHUL is able to outperform all the approaches relying on only GUI screenshots in almost every sub-category. Compared to other baselines we surpass them in Element accuracy and Step success rate, while remaining competitive in Operational F1. This indicates that the local elements detected by our HSP module and SEED descriptions provide highly valuable information for web navigation tasks. Although we provide better Operational F1 than HTML-based methods, we still falter when it comes to element accuracy and step success rate as predicting bounding boxes is a more complex task than selecting HTML elements.

\subsection{Screen Point-and-Read}
\label{sec:screenpnr}

\textbf{Dataset and experiments}- We use the Screen Point and Read\cite{fan2024readpointedlayoutawaregui} benchmark to evaluate TRISHUL's performance on the GUI referring task. It evaluates the accuracy of the generated content description \( \hat{D}_{\text{c}} \) and layout description \( \hat{D}_{\text{l}} \) for the region marked by the user over the interface. This benchmark comprises of 650 screenshots across three domains: web, mobile, and operating systems. To validate our method, we run experiments using GPT-4o \cite{gpt4o}, GPT-4V \cite{GPT-4V}, and Claude-3.5-Sonnet \cite{claude3.5}, enabling us to examine performance across multiple LVLMs.

\textbf{Evaluation and Results} - To assess the quality of the generated content description and layout description we employ the cycle consistency evaluation following the screen point-and-read \cite{fan2024readpointedlayoutawaregui} paper. The agent outputs (\( \hat{D}_{\text{c}} \) , \( \hat{D}_{\text{l}} \)) are fed into an auxiliary model, which is asked to complete a downstream task, with its performance indicating description quality. We benchmark our approach against baseline GPT-4o, Claude, and the ToL agent from Screen point-and-read, using GPT-4o, GPT-4V , and Claude-3.5-Sonnet as the primary models. We also compute language similarity metrics like BERT \cite{Zhang2019BERTScoreET} score and ROUGE-L \cite{Rouge} to evaluate alignment with human-verified ground truth.

To further validate quality, we conduct two rounds of human evaluation: the first compares our approach against baseline GPT-4o, while the second compares our approach with the ToL agent, both using GPT-4o as the primary LLM. We employ 10 human annotators from \cite{Indika} and ask them to choose between the description generated by our approach and the alternative approach. Each evaluator is presented with the labeled image and asked a single question \textit{“Given the image with the labeled point, which description do you prefer?”}.The majority vote is used to select the preferred description. To ensure unbiased evaluation the annotators are unaware of which model generates which descriptions. The annotators are compensated at minimum wage. 

\begin{figure}
    \centering
    \includegraphics[width=1\linewidth]{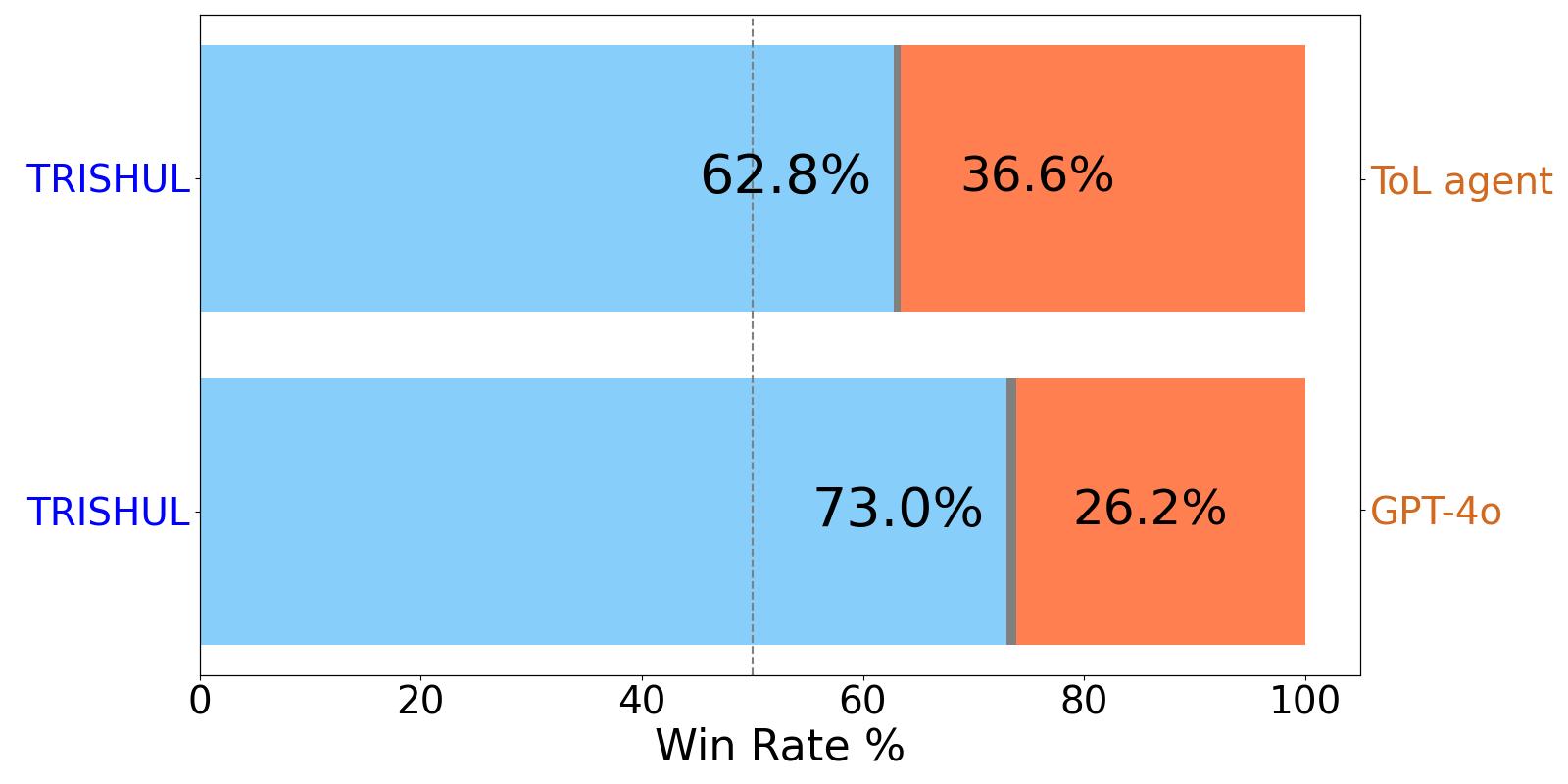}
    \caption{Human evaluation results on ScreenPR benchmark. TRISHUL is preferred by human annotators 63\% of the time over ToL agent and 73\% of the time over baseline GPT-4o}
    \label{fig:human_eval}
\end{figure}

TRISHUL consistently outperforms both the baseline and the ToL agent across all evaluation metrics for GPT-4V, Claude, and GPT-4o models (Table \ref{tab:screenpr}). Human evaluation results (Figure \ref{fig:human_eval}) further validate TRISHUL's efficacy, with descriptions generated by TRISHUL being preferred by annotators 73\% of the time over GPT-4o and 62.8\% of the time over ToL. TRISHUL ties with GPT-4o 0.9\% of the times and with ToL agent 0.6\% of the times.

\begin{figure}
    \centering
    
        \centering
        \includegraphics[width=\linewidth]{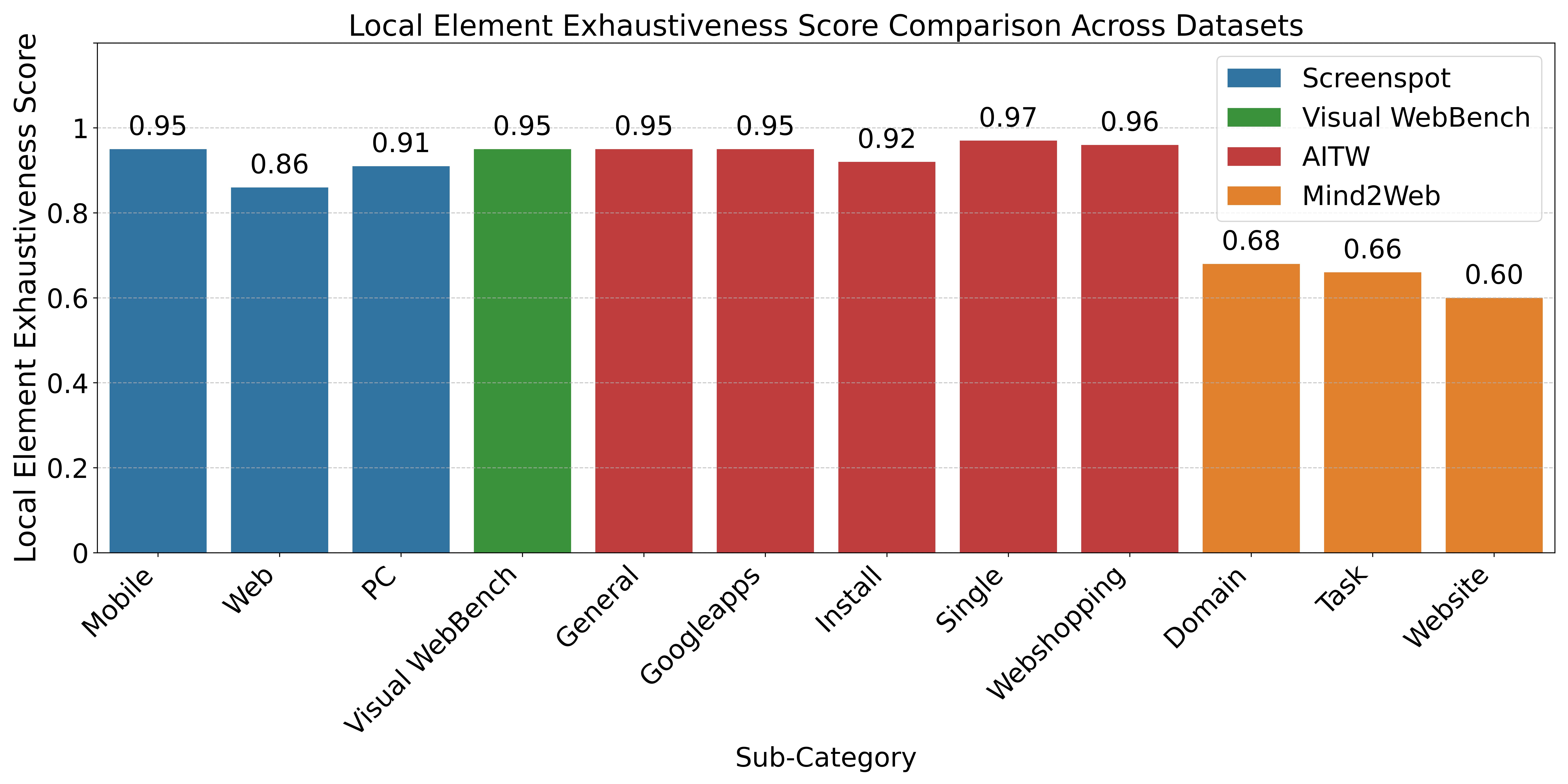}
        \caption{Local Element Exhaustiveness Score for ScreenSpot, Visual WebBench, AITW and Mind2Web}
        \label{fig:LEE}
\end{figure}

\section{Discussion}

\begin{table}[h!]
\small
    \centering
    \begin{tabular}{lcccccc}
        \toprule
        \multirow{2}{*}{Benchmark} & \multirow{2}{*}{Model} & \multicolumn{3}{c}{Accuracy (\%)} \\
        \cmidrule(lr){3-5}
        & & Pass@1 & Pass@2 & Pass@3 \\
        \midrule
        \multirow{3}{*}{VisualWebBench} & GPT-4o & 68.0 & 81.6 & 83.5 \\
                                        & GPT-4V & 56.3 & 69.9 & 71.8 \\
                                        & SeeClick & 31.0 & 36.0 & 36.0 \\
        \midrule
        \multirow{3}{*}{ScreenSpot} & GPT-4o & 72.2 & 77.8 & 80.0 \\
                                     & GPT-4V & 59.0 & 67.2 & 70.6 \\
                                     & SeeClick & 55.0 & 55.0 & 59.0 \\
        \bottomrule
    \end{tabular}
    \caption{Pass@1, Pass@2, and Pass@3 Accuracy (\%) for VisualWebBench and ScreenSpot using GPT-4o, GPT-4V,(with the TRISHUL framework) and SeeClick models.}
    \label{tab:pass_accuracy}
\end{table}

\subsection{Analysis on sampling multiple candidates}
LVLM-based GUI agents that rely solely on visual perception aim to mirror human like interface interaction. Humans often explore multiple paths when interacting with novel/complicated GUIs. Traditional metrics like pass@1 (top@1), may not fully reflect an agent's success in tasks that benefit from exploration. Recent research \cite{koh2024treesearchlanguagemodel}, shows that sampling and evaluating multiple potential action paths, then filtering them with a value model, improves success rates by reducing decision uncertainty.

The ToL agent has proven effective as a verification layer for mobile agents \cite{fan2024readpointedlayoutawaregui}, accurately identifying correct and incorrect action paths. Leveraging this insight, we propose utilizing TRISHUL as a verification agent in a GUI agent system to enable multi-click grounding with enhanced accuracy. Our findings in Table \ref{tab:pass_accuracy} indicate that multi-sampling metrics like pass@2 and pass@3 improve grounding accuracy by over 10\% across models on tasks in the ScreenSpot and VisualWebBench datasets. Here, pass@k highlights top K action-grounding candidates generated by TRISHUL. 



\subsection{Failure Analysis}

In Figure \ref{fig:LEE}, we evaluate the Local Element Exhaustiveness (LEE) metric across various datasets and their splits. The LEE score for an image is binary: it is set to 1 if the midpoint of the ground truth (GT) bounding box falls within any bounding box of local elements detected by our Hierarchical Screen Parsing (HSP) module; otherwise, it is set to 0. Thus, the LEE score shows the bottleneck that happens in our pipeline after the LE detection stage.

The results show a clear correlation between LEE scores and TRISHUL's performance across datasets. In particular, the low LEE scores in the Mind2Web dataset highlight that the limited exhaustiveness of local elements detected by the HSP module is a key factor constraining TRISHUL's effectiveness in web navigation tasks.

\section{Conclusion}
In this paper, we introduced TRISHUL, a training-free agentic framework that enables LVLMs to achieve comprehensive GUI screen understanding using two key modules: HSP and SEED. The HSP module organizes GUI elements into a multi-granular hierarchical structure, distinguishing Global Regions of Interest (GROIs) from local elements, while the SEED module enhances spatial context-aware reasoning. Experiments on ScreenSpot, VisualWebBench, AITW, Mind2Web, and ScreenPR demonstrate that TRISHUL outperforms all training-free methods and rivals training-based approaches while maintaining superior cross-task and cross-platform generalizability.

\section*{Impact Statement}

This work advances machine learning methods for comprehensive graphical user interface (GUI) comprehension, enabling more intuitive and automated interactions. A key positive impact lies in enhancing accessibility for visually challenged individuals through our GUI referring agent, helping them navigate digital environments more effectively. Potential ethical concerns primarily revolve around privacy and control, if such automation tools are misused. Overall, while this framework promises to streamline user experiences and empower those with visual impairments, continued vigilance is advised to safeguard responsible, transparent, and privacy-oriented deployment.

\bibliography{MAIN}

\begin{thebibliography}{47}
\providecommand{\natexlab}[1]{#1}
\providecommand{\url}[1]{\texttt{#1}}
\expandafter\ifx\csname urlstyle\endcsname\relax
  \providecommand{\doi}[1]{doi: #1}\else
  \providecommand{\doi}{doi: \begingroup \urlstyle{rm}\Url}\fi

\bibitem[Anthropic(2023)]{claude3.5}
Anthropic.
\newblock Introducing claude 3.5, 2023.
\newblock URL \url{https://www-cdn.anthropic.com/fed9cc193a14b84131812372d8d5857f8f304c52/Model_Card_Claude_3_Addendum.pdf}.

\bibitem[Bai et~al.(2021)Bai, Zang, Xu, Sunkara, Rastogi, Chen, and y~Arcas]{Bai2021UIBertLG}
Bai, C., Zang, X., Xu, Y., Sunkara, S., Rastogi, A., Chen, J., and y~Arcas, B.~A.
\newblock Uibert: Learning generic multimodal representations for ui understanding.
\newblock In \emph{International Joint Conference on Artificial Intelligence}, 2021.
\newblock URL \url{https://api.semanticscholar.org/CorpusID:236493482}.

\bibitem[Bai et~al.(2024)Bai, Zhou, Cemri, Pan, Suhr, Levine, and Kumar]{Bai2024DigiRLTI}
Bai, H., Zhou, Y., Cemri, M., Pan, J., Suhr, A., Levine, S., and Kumar, A.
\newblock Digirl: Training in-the-wild device-control agents with autonomous reinforcement learning.
\newblock \emph{ArXiv}, abs/2406.11896, 2024.
\newblock URL \url{https://api.semanticscholar.org/CorpusID:270562229}.

\bibitem[Brown et~al.(2020)Brown, Mann, Ryder, Subbiah, Kaplan, Dhariwal, Neelakantan, Shyam, Sastry, Askell, Agarwal, Herbert-Voss, Krueger, Henighan, Child, Ramesh, Ziegler, Wu, Winter, Hesse, Chen, Sigler, teusz Litwin, Gray, Chess, Clark, Berner, McCandlish, Radford, Sutskever, and Amodei]{Brown2020LanguageMA}
Brown, T.~B., Mann, B., Ryder, N., Subbiah, M., Kaplan, J., Dhariwal, P., Neelakantan, A., Shyam, P., Sastry, G., Askell, A., Agarwal, S., Herbert-Voss, A., Krueger, G., Henighan, T., Child, R., Ramesh, A., Ziegler, D.~M., Wu, J., Winter, C., Hesse, C., Chen, M., Sigler, E., teusz Litwin, M., Gray, S., Chess, B., Clark, J., Berner, C., McCandlish, S., Radford, A., Sutskever, I., and Amodei, D.
\newblock Language models are few-shot learners.
\newblock \emph{ArXiv}, abs/2005.14165, 2020.
\newblock URL \url{https://api.semanticscholar.org/CorpusID:218971783}.

\bibitem[Chen et~al.(2020{\natexlab{a}})Chen, Chen, Xing, Xu, Zhu, Li, and Wang]{Chen2020UnblindYA}
Chen, J., Chen, C., Xing, Z., Xu, X., Zhu, L., Li, G., and Wang, J.
\newblock Unblind your apps: Predicting natural-language labels for mobile gui components by deep learning.
\newblock \emph{2020 IEEE/ACM 42nd International Conference on Software Engineering (ICSE)}, pp.\  322--334, 2020{\natexlab{a}}.
\newblock URL \url{https://api.semanticscholar.org/CorpusID:211677644}.

\bibitem[Chen et~al.(2020{\natexlab{b}})Chen, Xie, Xing, Chen, Xu, Zhu, and Li]{Chen2020ObjectDF}
Chen, J., Xie, M., Xing, Z., Chen, C., Xu, X., Zhu, L., and Li, G.
\newblock Object detection for graphical user interface: old fashioned or deep learning or a combination?
\newblock \emph{Proceedings of the 28th ACM Joint Meeting on European Software Engineering Conference and Symposium on the Foundations of Software Engineering}, 2020{\natexlab{b}}.
\newblock URL \url{https://api.semanticscholar.org/CorpusID:221103890}.

\bibitem[Cheng et~al.(2024)Cheng, Sun, Chu, Xu, Li, Zhang, and Wu]{Cheng2024SeeClickHG}
Cheng, K., Sun, Q., Chu, Y., Xu, F., Li, Y., Zhang, J., and Wu, Z.
\newblock Seeclick: Harnessing gui grounding for advanced visual gui agents.
\newblock In \emph{Annual Meeting of the Association for Computational Linguistics}, 2024.
\newblock URL \url{https://api.semanticscholar.org/CorpusID:267069082}.

\bibitem[Deng et~al.(2023)Deng, Gu, Zheng, Chen, Stevens, Wang, Sun, and Su]{Deng2023Mind2WebTA}
Deng, X., Gu, Y., Zheng, B., Chen, S., Stevens, S., Wang, B., Sun, H., and Su, Y.
\newblock Mind2web: Towards a generalist agent for the web.
\newblock \emph{ArXiv}, abs/2306.06070, 2023.
\newblock URL \url{https://api.semanticscholar.org/CorpusID:259129428}.

\bibitem[Fan et~al.(2024)Fan, Ding, Kuo, Jiang, Zhao, Guan, Yang, Zhang, and Wang]{fan2024readpointedlayoutawaregui}
Fan, Y., Ding, L., Kuo, C.-C., Jiang, S., Zhao, Y., Guan, X., Yang, J., Zhang, Y., and Wang, X.~E.
\newblock Read anywhere pointed: Layout-aware gui screen reading with tree-of-lens grounding, 2024.
\newblock URL \url{https://arxiv.org/abs/2406.19263}.

\bibitem[Furuta et~al.(2023)Furuta, Nachum, Lee, Matsuo, Gu, and Gur]{Furuta2023MultimodalWN}
Furuta, H., Nachum, O., Lee, K.-H., Matsuo, Y., Gu, S.~S., and Gur, I.
\newblock Multimodal web navigation with instruction-finetuned foundation models.
\newblock \emph{ArXiv}, abs/2305.11854, 2023.
\newblock URL \url{https://api.semanticscholar.org/CorpusID:258823350}.

\bibitem[Gur et~al.(2018)Gur, R{\"u}ckert, Faust, and Hakkani-T{\"u}r]{Gur2018LearningTN}
Gur, I., R{\"u}ckert, U., Faust, A., and Hakkani-T{\"u}r, D.~Z.
\newblock Learning to navigate the web.
\newblock \emph{ArXiv}, abs/1812.09195, 2018.
\newblock URL \url{https://api.semanticscholar.org/CorpusID:56657805}.

\bibitem[Gur et~al.(2023)Gur, Furuta, Huang, Safdari, Matsuo, Eck, and Faust]{Gur2023ARW}
Gur, I., Furuta, H., Huang, A., Safdari, M., Matsuo, Y., Eck, D., and Faust, A.
\newblock A real-world webagent with planning, long context understanding, and program synthesis.
\newblock \emph{ArXiv}, abs/2307.12856, 2023.
\newblock URL \url{https://api.semanticscholar.org/CorpusID:260126067}.

\bibitem[He et~al.(2024)He, Yao, Ma, Yu, Dai, Zhang, Lan, and Yu]{He2024WebVoyagerBA}
He, H., Yao, W., Ma, K., Yu, W., Dai, Y., Zhang, H., Lan, Z., and Yu, D.
\newblock Webvoyager: Building an end-to-end web agent with large multimodal models.
\newblock In \emph{Annual Meeting of the Association for Computational Linguistics}, 2024.
\newblock URL \url{https://api.semanticscholar.org/CorpusID:267211622}.

\bibitem[He et~al.(2020)He, Sunkara, Zang, Xu, Liu, Wichers, Schubiner, Lee, and Chen]{He2020ActionBertLU}
He, Z., Sunkara, S., Zang, X., Xu, Y., Liu, L., Wichers, N., Schubiner, G., Lee, R.~B., and Chen, J.
\newblock Actionbert: Leveraging user actions for semantic understanding of user interfaces.
\newblock In \emph{AAAI Conference on Artificial Intelligence}, 2020.
\newblock URL \url{https://api.semanticscholar.org/CorpusID:229363676}.

\bibitem[Hong et~al.(2023)Hong, Wang, Lv, Xu, Yu, Ji, Wang, Wang, Zhang, Li, Xu, Dong, Ding, and Tang]{Hong2023CogAgentAV}
Hong, W., Wang, W., Lv, Q., Xu, J., Yu, W., Ji, J., Wang, Y., Wang, Z., Zhang, Y., Li, J.-Z., Xu, B., Dong, Y., Ding, M., and Tang, J.
\newblock Cogagent: A visual language model for gui agents.
\newblock \emph{ArXiv}, abs/2312.08914, 2023.
\newblock URL \url{https://api.semanticscholar.org/CorpusID:273102270}.

\bibitem[https://www.indikaai.com/()]{Indika}
https://www.indikaai.com/.
\newblock \emph{Indika.ai}.

\bibitem[JaidedAI()]{EasyOCR}
JaidedAI.
\newblock Easyocr: Ready-to-use ocr with 80+ supported languages and all popular writing scripts including latin, chinese, arabic, devanagari, cyrillic and etc.
\newblock URL \url{https://github.com/JaidedAI/EasyOCR}.

\bibitem[Jocher et~al.(2023)Jocher, Chaurasia, and Qiu]{yolov8_ultralytics}
Jocher, G., Chaurasia, A., and Qiu, J.
\newblock Ultralytics yolov8, 2023.
\newblock URL \url{https://github.com/ultralytics/ultralytics}.

\bibitem[Jurmu et~al.(2008)Jurmu, Boring, and Riekki]{Jurmu2008ScreenSpotMR}
Jurmu, M., Boring, S., and Riekki, J.
\newblock Screenspot: multidimensional resource discovery for distributed applications in smart spaces.
\newblock In \emph{International Conference on Mobile and Ubiquitous Systems: Networking and Services}, 2008.
\newblock URL \url{https://api.semanticscholar.org/CorpusID:192633}.

\bibitem[Kirillov et~al.(2023)Kirillov, Mintun, Ravi, Mao, Rolland, Gustafson, Xiao, Whitehead, Berg, Lo, Doll{\'a}r, and Girshick]{SAM}
Kirillov, A., Mintun, E., Ravi, N., Mao, H., Rolland, C., Gustafson, L., Xiao, T., Whitehead, S., Berg, A.~C., Lo, W.-Y., Doll{\'a}r, P., and Girshick, R.
\newblock Segment anything.
\newblock \emph{arXiv:2304.02643}, 2023.

\bibitem[Koh et~al.(2024)Koh, McAleer, Fried, and Salakhutdinov]{koh2024treesearchlanguagemodel}
Koh, J.~Y., McAleer, S., Fried, D., and Salakhutdinov, R.
\newblock Tree search for language model agents, 2024.
\newblock URL \url{https://arxiv.org/abs/2407.01476}.

\bibitem[Li et~al.(2023)Li, Li, Savarese, and Hoi]{blip2}
Li, J., Li, D., Savarese, S., and Hoi, S.
\newblock Blip-2: Bootstrapping language-image pre-training with frozen image encoders and large language models.
\newblock \emph{arXiv preprint arXiv:2301.12597}, 2023.

\bibitem[Li et~al.(2020{\natexlab{a}})Li, He, Zhou, Zhang, and Baldridge]{Li2020MappingNL}
Li, Y., He, J., Zhou, X., Zhang, Y., and Baldridge, J.
\newblock Mapping natural language instructions to mobile ui action sequences.
\newblock \emph{ArXiv}, abs/2005.03776, 2020{\natexlab{a}}.
\newblock URL \url{https://api.semanticscholar.org/CorpusID:218571167}.

\bibitem[Li et~al.(2020{\natexlab{b}})Li, Li, He, Zheng, Li, and Guan]{Li2020WidgetCG}
Li, Y., Li, G., He, L., Zheng, J., Li, H., and Guan, Z.
\newblock Widget captioning: Generating natural language description for mobile user interface elements.
\newblock In \emph{Conference on Empirical Methods in Natural Language Processing}, 2020{\natexlab{b}}.
\newblock URL \url{https://api.semanticscholar.org/CorpusID:222272319}.

\bibitem[Lin(2004)]{Rouge}
Lin, C.-Y.
\newblock Rouge: A package for automatic evaluation of summaries.
\newblock pp.\ ~10, 01 2004.

\bibitem[Liu et~al.(2018)Liu, Guu, Pasupat, Shi, and Liang]{Liu2018ReinforcementLO}
Liu, E.~Z., Guu, K., Pasupat, P., Shi, T., and Liang, P.
\newblock Reinforcement learning on web interfaces using workflow-guided exploration.
\newblock \emph{ArXiv}, abs/1802.08802, 2018.
\newblock URL \url{https://api.semanticscholar.org/CorpusID:3530344}.

\bibitem[Liu et~al.(2024)Liu, Song, Lin, Lam, Neubig, Li, and Yue]{Liu2024VisualWebBenchHF}
Liu, J., Song, Y., Lin, B.~Y., Lam, W., Neubig, G., Li, Y., and Yue, X.
\newblock Visualwebbench: How far have multimodal llms evolved in web page understanding and grounding?
\newblock \emph{ArXiv}, abs/2404.05955, 2024.
\newblock URL \url{https://api.semanticscholar.org/CorpusID:269009925}.

\bibitem[Lu et~al.(2024)Lu, Yang, Shen, and Awadallah]{OmniParser}
Lu, Y., Yang, J., Shen, Y., and Awadallah, A.
\newblock Omniparser.
\newblock \emph{arXiv preprint arXiv:2408.00203}, 2024.

\bibitem[OpenAI(June, 2024{\natexlab{a}})]{GPT-4V}
OpenAI.
\newblock "gpt-4v(ision) system card", June, 2024{\natexlab{a}}.
\newblock URL \url{https://openai.com/index/gpt-4v-system-card/}.

\bibitem[OpenAI(June, 2024{\natexlab{b}})]{gpt4o}
OpenAI.
\newblock "hello gpt-4o.", June, 2024{\natexlab{b}}.
\newblock URL \url{https://openai.com/index/hello-gpt-4o/}.

\bibitem[Rawles et~al.(2023)Rawles, Li, Rodriguez, Riva, and Lillicrap]{rawles2023androidwildlargescaledataset}
Rawles, C., Li, A., Rodriguez, D., Riva, O., and Lillicrap, T.
\newblock Android in the wild: A large-scale dataset for android device control, 2023.
\newblock URL \url{https://arxiv.org/abs/2307.10088}.

\bibitem[Shaw et~al.(2023)Shaw, Joshi, Cohan, Berant, Pasupat, Hu, Khandelwal, Lee, and Toutanova]{shaw2023pixels}
Shaw, P., Joshi, M., Cohan, J., Berant, J., Pasupat, P., Hu, H., Khandelwal, U., Lee, K., and Toutanova, K.
\newblock From pixels to ui actions: Learning to follow instructions via graphical user interfaces.
\newblock In \emph{Advances in Neural Information Processing Systems}, 2023.
\newblock URL \url{https://arxiv.org/abs/2306.00245}.

\bibitem[Shi et~al.(2017)Shi, Karpathy, Fan, Hernandez, and Liang]{pmlr-v70-shi17a}
Shi, T., Karpathy, A., Fan, L., Hernandez, J., and Liang, P.
\newblock World of bits: An open-domain platform for web-based agents.
\newblock In Precup, D. and Teh, Y.~W. (eds.), \emph{Proceedings of the 34th International Conference on Machine Learning}, volume~70 of \emph{Proceedings of Machine Learning Research}, pp.\  3135--3144. PMLR, 06--11 Aug 2017.
\newblock URL \url{https://proceedings.mlr.press/v70/shi17a.html}.

\bibitem[Sridhar et~al.(2023)Sridhar, Lo, Xu, Zhu, and Zhou]{Sridhar2023HierarchicalPA}
Sridhar, A., Lo, R., Xu, F.~F., Zhu, H., and Zhou, S.
\newblock Hierarchical prompting assists large language model on web navigation.
\newblock In \emph{Conference on Empirical Methods in Natural Language Processing}, 2023.
\newblock URL \url{https://api.semanticscholar.org/CorpusID:258841249}.

\bibitem[Wang et~al.(2021)Wang, Li, Zhou, Chen, Grossman, and Li]{Wang2021Screen2WordsAM}
Wang, B., Li, G., Zhou, X., Chen, Z., Grossman, T., and Li, Y.
\newblock Screen2words: Automatic mobile ui summarization with multimodal learning.
\newblock \emph{The 34th Annual ACM Symposium on User Interface Software and Technology}, 2021.
\newblock URL \url{https://api.semanticscholar.org/CorpusID:236957064}.

\bibitem[Wei et~al.(2023)Wei, Wang, Schuurmans, Bosma, Ichter, Xia, Chi, Le, and Zhou]{wei2023chainofthoughtpromptingelicitsreasoning}
Wei, J., Wang, X., Schuurmans, D., Bosma, M., Ichter, B., Xia, F., Chi, E., Le, Q., and Zhou, D.
\newblock "chain-of-thought prompting elicits reasoning in large language models", 2023.
\newblock URL \url{https://arxiv.org/abs/2201.11903}.

\bibitem[Wu et~al.(2021)Wu, Zhang, Nichols, and Bigham]{Wu2021ScreenPT}
Wu, J., Zhang, X., Nichols, J., and Bigham, J.~P.
\newblock Screen parsing: Towards reverse engineering of ui models from screenshots.
\newblock \emph{The 34th Annual ACM Symposium on User Interface Software and Technology}, 2021.
\newblock URL \url{https://api.semanticscholar.org/CorpusID:237571719}.

\bibitem[Xie et~al.(2024)Xie, Zhang, Chen, Li, Zhao, Cao, Hua, Cheng, Shin, Lei, Liu, Xu, Zhou, Savarese, Xiong, Zhong, and Yu]{Xie2024OSWorldBM}
Xie, T., Zhang, D., Chen, J., Li, X., Zhao, S., Cao, R., Hua, T.~J., Cheng, Z., Shin, D., Lei, F., Liu, Y., Xu, Y., Zhou, S., Savarese, S., Xiong, C., Zhong, V., and Yu, T.
\newblock Osworld: Benchmarking multimodal agents for open-ended tasks in real computer environments.
\newblock \emph{ArXiv}, abs/2404.07972, 2024.
\newblock URL \url{https://api.semanticscholar.org/CorpusID:269042918}.

\bibitem[Yan et~al.(2023)Yan, Yang, Zhu, Lin, Li, Wang, Yang, Zhong, McAuley, Gao, Liu, and Wang]{Yan2023GPT4VIW}
Yan, A., Yang, Z., Zhu, W., Lin, K.~Q., Li, L., Wang, J., Yang, J., Zhong, Y., McAuley, J.~J., Gao, J., Liu, Z., and Wang, L.
\newblock Gpt-4v in wonderland: Large multimodal models for zero-shot smartphone gui navigation.
\newblock \emph{ArXiv}, abs/2311.07562, 2023.
\newblock URL \url{https://api.semanticscholar.org/CorpusID:265149992}.

\bibitem[Yang et~al.(2023)Yang, Zhang, Li, Zou, Li, and Gao]{yang2023setofmark}
Yang, J., Zhang, H., Li, F., Zou, X., Li, C., and Gao, J.
\newblock Set-of-mark prompting unleashes extraordinary visual grounding in gpt-4v.
\newblock \emph{arXiv preprint arXiv:2310.11441}, 2023.

\bibitem[Yao et~al.(2022)Yao, Chen, Yang, and Narasimhan]{Yao2022WebShopTS}
Yao, S., Chen, H., Yang, J., and Narasimhan, K.
\newblock Webshop: Towards scalable real-world web interaction with grounded language agents.
\newblock \emph{ArXiv}, abs/2207.01206, 2022.
\newblock URL \url{https://api.semanticscholar.org/CorpusID:250264533}.

\bibitem[You et~al.(2024)You, Zhang, Schoop, Weers, Swearngin, Nichols, Yang, and Gan]{You2024FerretUIGM}
You, K., Zhang, H., Schoop, E., Weers, F., Swearngin, A., Nichols, J., Yang, Y., and Gan, Z.
\newblock Ferret-ui: Grounded mobile ui understanding with multimodal llms.
\newblock In \emph{European Conference on Computer Vision}, 2024.
\newblock URL \url{https://api.semanticscholar.org/CorpusID:269005503}.

\bibitem[Zhang et~al.(2023)Zhang, Yang, Liu, Han, Chen, Huang, Fu, and Yu]{Zhang2023AppAgentMA}
Zhang, C.~X., Yang, Z., Liu, J., Han, Y., Chen, X., Huang, Z., Fu, B., and Yu, G.
\newblock Appagent: Multimodal agents as smartphone users.
\newblock \emph{ArXiv}, abs/2312.13771, 2023.
\newblock URL \url{https://api.semanticscholar.org/CorpusID:266435868}.

\bibitem[Zhang et~al.(2019)Zhang, Kishore, Wu, Weinberger, and Artzi]{Zhang2019BERTScoreET}
Zhang, T., Kishore, V., Wu, F., Weinberger, K.~Q., and Artzi, Y.
\newblock Bertscore: Evaluating text generation with bert.
\newblock \emph{ArXiv}, abs/1904.09675, 2019.
\newblock URL \url{https://api.semanticscholar.org/CorpusID:127986044}.

\bibitem[Zhang et~al.(2021)Zhang, de~Greef, Swearngin, White, Murray, Yu, Shan, Nichols, Wu, Fleizach, Everitt, and Bigham]{Zhang2021ScreenRC}
Zhang, X., de~Greef, L., Swearngin, A., White, S., Murray, K.~I., Yu, L., Shan, Q., Nichols, J., Wu, J., Fleizach, C., Everitt, A., and Bigham, J.~P.
\newblock Screen recognition: Creating accessibility metadata for mobile applications from pixels.
\newblock \emph{Proceedings of the 2021 CHI Conference on Human Factors in Computing Systems}, 2021.
\newblock URL \url{https://api.semanticscholar.org/CorpusID:231592643}.

\bibitem[Zheng et~al.(2024)Zheng, Gou, Kil, Sun, and Su]{zheng2023seeact}
Zheng, B., Gou, B., Kil, J., Sun, H., and Su, Y.
\newblock Gpt-4v(ision) is a generalist web agent, if grounded.
\newblock \emph{arXiv preprint arXiv:2401.01614}, 2024.

\bibitem[Zhou et~al.(2023)Zhou, Xu, Zhu, Zhou, Lo, Sridhar, Cheng, Bisk, Fried, Alon, et~al.]{zhou2023webarena}
Zhou, S., Xu, F.~F., Zhu, H., Zhou, X., Lo, R., Sridhar, A., Cheng, X., Bisk, Y., Fried, D., Alon, U., et~al.
\newblock Webarena: A realistic web environment for building autonomous agents.
\newblock \emph{arXiv preprint arXiv:2307.13854}, 2023.
\newblock URL \url{https://webarena.dev}.

\end{thebibliography}
\bibliographystyle{icml2025}

\newpage
\appendix
\onecolumn
\section{Appendix.}
\label{sec:appendix}

\subsection{Model Specifications and Endpoints}
\label{app:subsection_label}
Since all our work leverages closed-source models like GPT-4V, GPT-4o, and Claude, we mention the model identifiers that we use for our API calls for clarity. For GPT-4V - "gpt-4-vision-preview", For GPT-4o - "gpt-4o-2024-08-06", and for Claude - "claude-3-5-sonnet-20241022". Unless otherwise noted, all experiments are conducted with a temperature setting of 0.0.

\subsection{Hierarchical Screen Parsing Details}
\label{app: HSP_details}

\subsubsection{IoS Score} Similar to IoU score we define an IoS score as:
\lstset{
  language=Python,
  backgroundcolor=\color{white},
  basicstyle=\ttfamily\small,
  breaklines=true,
  frame=single,
  keywordstyle=\color{blue}\bfseries,
  commentstyle=\color{green!60!black},
  stringstyle=\color{red},
  showstringspaces=false,
}
\begin{lstlisting}
def IoS(boxA, boxB):
    xA = max(boxA[0], boxB[0])
    yA = max(boxA[1], boxB[1])
    xB = min(boxA[2], boxB[2])
    yB = min(boxA[3], boxB[3])
    interArea = max(0, xB - xA) * max(0, yB - yA)
    boxAArea = (boxA[2] - boxA[0]) * (boxA[3] - boxA[1])
    ios = interArea / float(boxAArea + 1e-3)
    return ios
\end{lstlisting}
The IoS (Intersection over Size) score is a measure used to evaluate the overlap between two bounding boxes, typically in the context of object detection. It calculates the ratio of the intersection area between two boxes to the area of the first box. IoS(A, B) (also written as $IOS_{A}$), a score of 0.5 means 50\% of A intersects with B.

\subsubsection{Filtering Redundant Bounding boxes}
\label{sec:filtering}
\begin{figure*}
    \centering
    \includegraphics[width=1\linewidth]{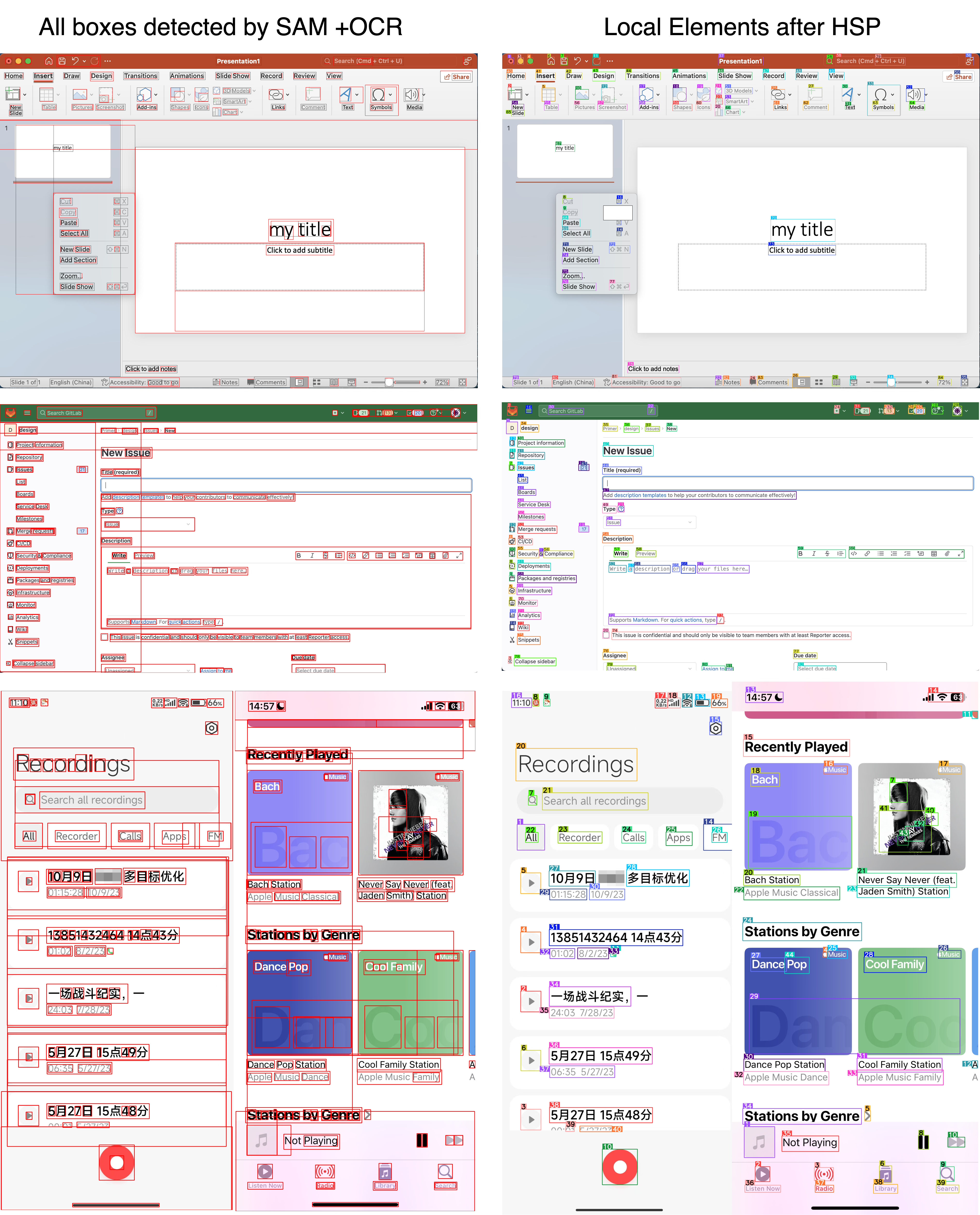}
    \caption{Candidate bounding boxes generated from SAM + OCR to the left and the corresponding  HSP results (Icon + text + picture) to the right}
    \label{fig:candidate_ex}
\end{figure*}

The output of EasyOCR + SAM model combined is extremely cluttered (see \ref{fig:candidate_ex}) and contains numerous overlaps and false positive detections from both models. We deploy the following steps to parse the outputs of SAM and OCR (local elements as referred to in the main script) together. 
\begin{itemize}

    \item \textbf{Generate  GROI, Icon and Button Candidate Proposals} 
    Classify all SAM boxes based on $A_{thresh-GROI}$, $A_{thresh-icon}$, $A_{thresh-button}$. Let \( B \) represent the set of bounding boxes detected in the GUI. 

    Global Region of Interest (GROI) Candidates: The set of boxes with an area greater than the GROI threshold is given by:
       \[
       \text{GROI} = \{ b \in B \mid \text{Area}(b) > A_{thresh-GROI} \}
       \]
    
    Icon Candidates:The set of boxes with an area between the Button and Icon thresholds is defined as:
       \[
       \text{Icon} = \{ b \in B \mid A_{thresh-button} < \text{Area}(b) < A_{thresh-icon} \}
       \]
    
    Button Candidates: The set of boxes with an area less than the Button threshold is:
       \[
       \text{Button} = \{ b \in B \mid \text{Area}(b) < A_{thresh-button} \}
       \]

   \item \textbf{Remove False Positive Text Bounding Boxes:} 
    Remove text boxes using a predefined dictionary,  that are likely OCR mis-detections for icons. These texts usually contain  only special characters or short, meaningless words. If a word contains one of these characters and has a length of less than < 5 that text bbox is ignored.
    
    Characters/Words to ignore:
      
    \begin{itemize}
        \item \texttt{"@", "\#", "x", "?", "\{", "\}", "<", ">", "\&", "`", "\textasciitilde{}", "\\", "=", "C", "Q", "88", "83", "98", "15J", "\textasciicircum{}", "0e", "n", "E", "ya", "ch", "893"}
    \end{itemize}

    \item \textbf{Remove Icons Inside or Overlapping Text Bounding Boxes:} 
    Remove icon bounding boxes that are either inside or intersect with text boxes, as they are likely to be text misidentified as icons by SAM.

    \item \textbf{Filter Square-like Icon Bounding Boxes:} 
    Keep only icons that are roughly square-shaped, based on a specific aspect ratio range of [0.7, 1.3].

    \item \textbf{Remove Redundant Icon and Button Bounding Boxes:} 
    Remove icon bounding boxes that are redundant, i.e., those that are inside or significantly overlap with $ IoS > 0.6$ with other icons or text boxes.

\end{itemize}

\subsubsection{Non Max Suppression for GROIs}
\begin{itemize}
     \item \textbf{Reject  boxes with low Information score$S$} 
    If the current bounding box has an information  score $S < S_{thresh}$ it is rejected. $S_{thresh}$ is set to 10 for the ScreenPoint and Read task and 25 for action grounding task.
    
    \item \textbf{Reject Overlapping BBoxes:} 
    If the current bounding box intersects with a previously selected bounding box with a higher Infrormation score and $IoS_{current} > IoS_{overlap-thresh}$ it is rejected. $IoS_{overlap-thresh}$ thresh is set to 0.5 for visual grounding task and 0 for ScreenPR task.

    \item \textbf{Reject Contained BBoxes (Smaller GROIs Inside Larger):} 
    If the current bounding box is inside a previously selected bounding box with a higher Information Score and if $IoS_{current} > IoS_{inside-thresh}$ it is rejected. $IoS_{overlap-thresh}$ thresh is set to 0.5 for visual grounding task and 0 for ScreenPR task.

    \item \textbf{Reject Engulfing Bboxes (Larger GROIs Inside Smaller):} 
    If the current bounding box completely engulfs a bounding box with a higher Information Score then it is rejected.
    
\end{itemize}

\subsection{GROI analysis for ScreenSpot and VisualWebBench}

\begin{figure*}
    \centering
    \includegraphics[width=1\linewidth]{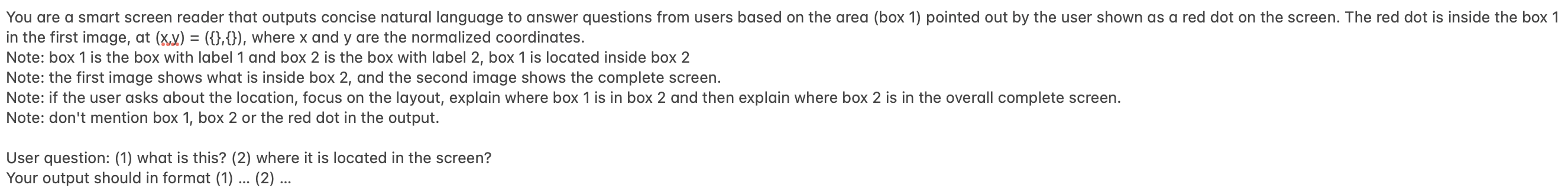}
    \caption{Prompt for Screen Point-and-Read}
    \label{fig:prompt_screenpnr}
\end{figure*}

\begin{figure*}
    \begin{minipage}{0.45\textwidth}
        \centering
        \includegraphics[width=\textwidth]{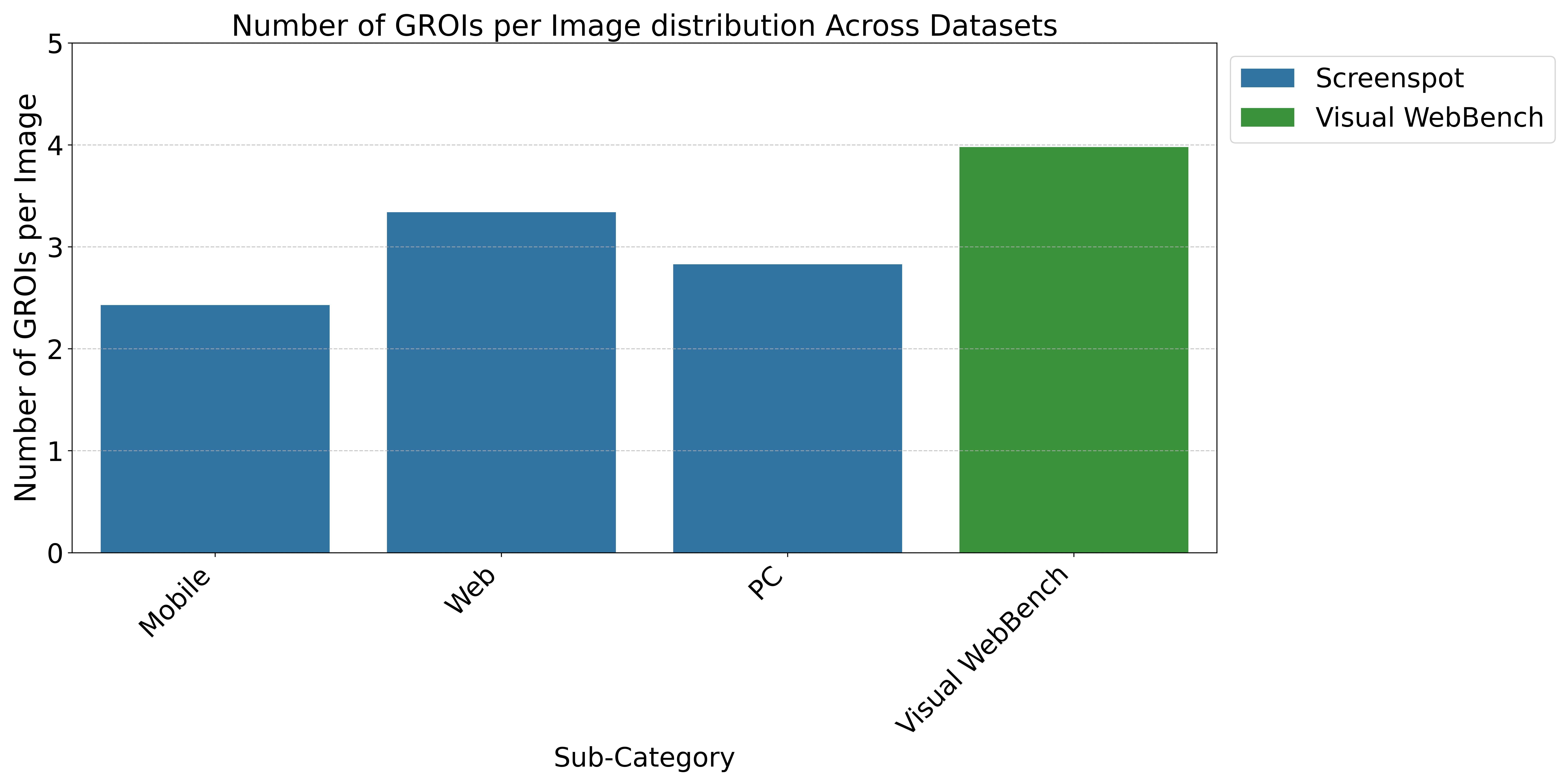}
        \caption{Distribution of Number of GROIs per image for ScreenSpot and Visual WebBench}
        \label{fig:image1}
    \end{minipage}%
    \hfill
    \begin{minipage}{0.45\textwidth}
        \centering
        \includegraphics[width=\textwidth]{main_figs/fig5.png}
        \caption{Distribution of Total GROI area / Image area for ScreenSpot and Visual WebBench}
        \label{fig:image2}
    \end{minipage}
\end{figure*}

\label{sec6.3}
We plot two additional statistics for the detected GROI's through our HSP block. In Figure \ref{fig:image1}
we plot the average number of GROIs per image, across the three different sub-categories of ScreenSpot and the full dataset of VisualWebBench. We observe that GUI screenshots from mobiles have the lowest average number of GROIs per image. This is due to the fact that mobile regions are not semantically coherent, therefore lesser number of GROIs are generated.

In Figure \ref{fig:image2} we plot the average of the total area covered by all GROIs in an image to the total area of the image. Mobile GUI screenshots have the least dense GROI coverage, due to the fact that we also detect fewer GROIs in mobile screenshots. These studies further validate the fact that GROIs are not as useful for mobile GUI's however they offer more benefit for PC and web based GUIs. 
\begin{figure*}
    \centering
    \includegraphics[width=1\linewidth]{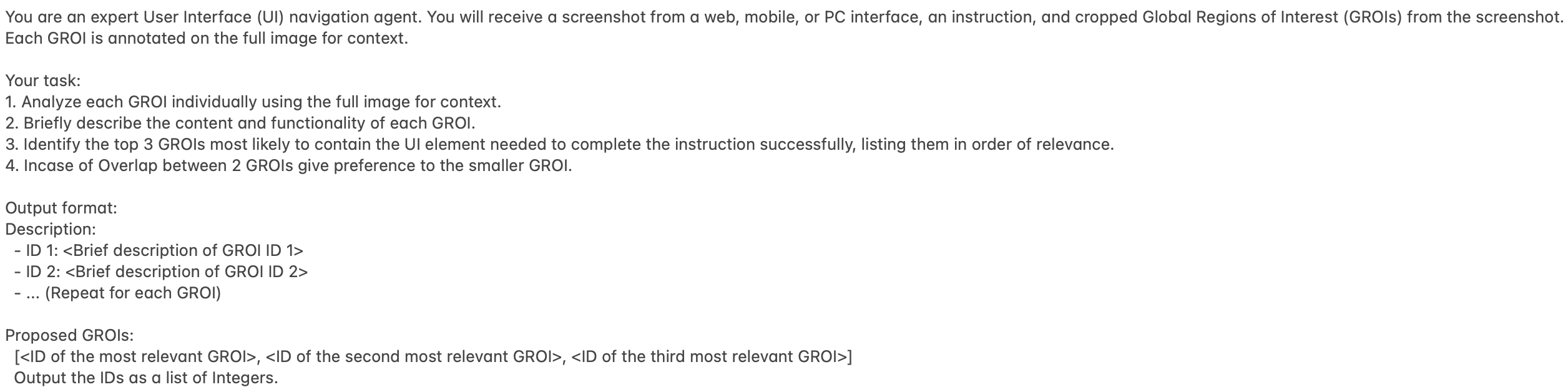}
    \caption{Prompt for instruction guided GROI Proposal generation}
    \label{fig:prompt_groi}
\end{figure*}

\begin{figure*}
    \centering
    \includegraphics[width=1\linewidth]{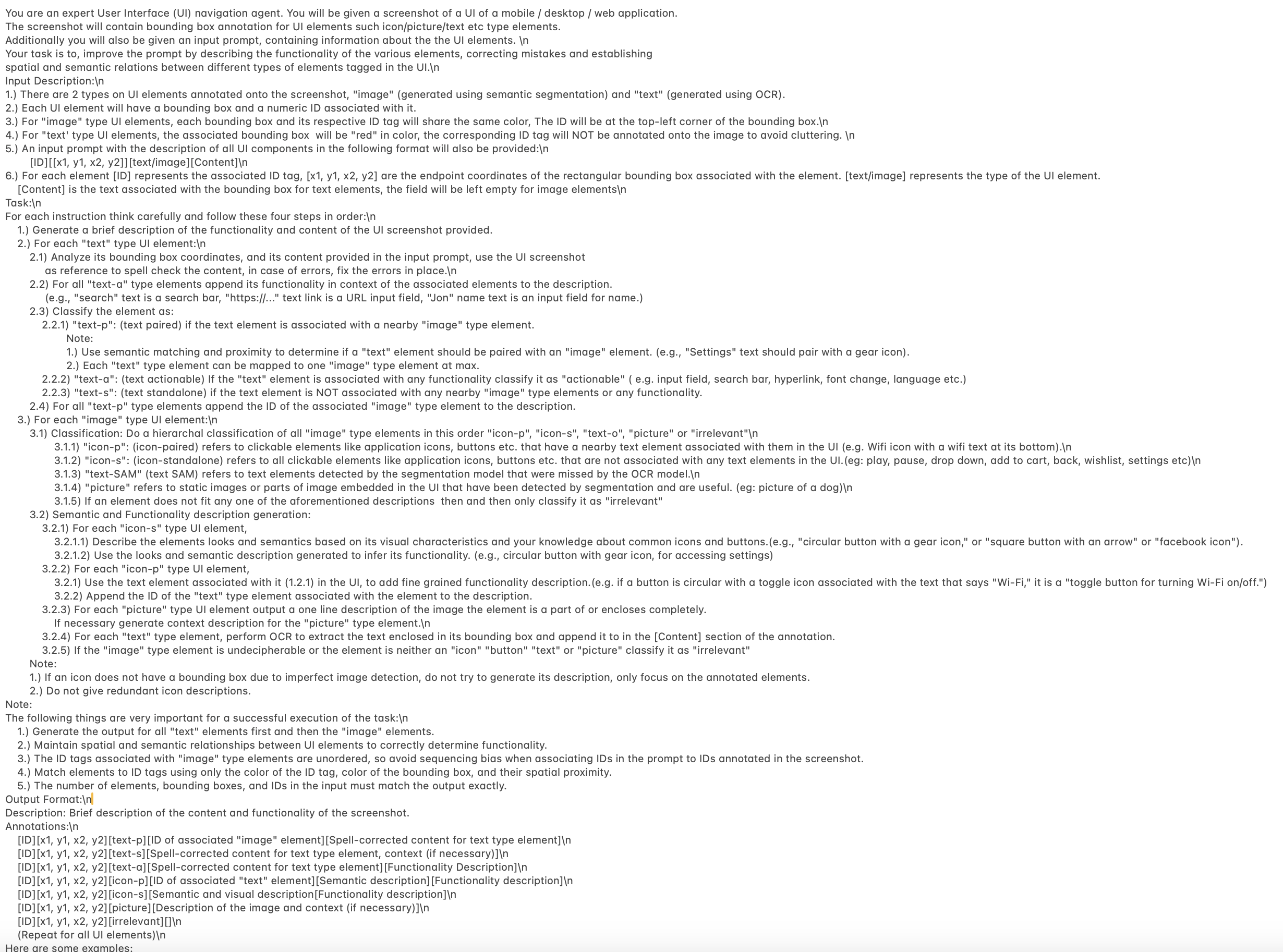}
    \caption{Prompt for SEED}
    \label{fig:prompt_seed}
\end{figure*}

\begin{figure*}
    \centering
    \includegraphics[width=1\linewidth]{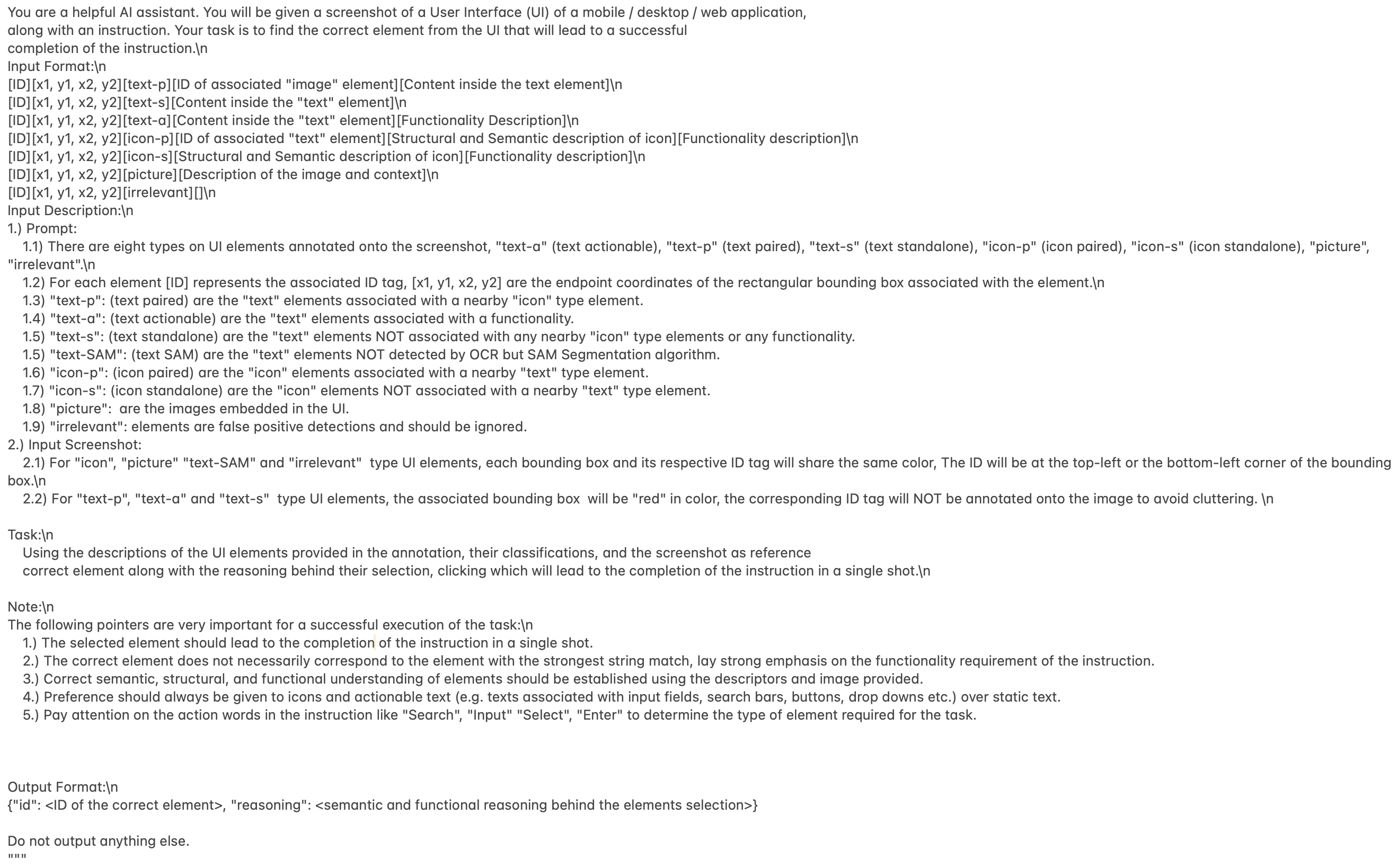}
    \caption{SoM grounding Prompt for ScreenSpot and VisualWebBench}
    \label{fig:prompt_screenspot}
\end{figure*}

\begin{figure*}
    \centering
    \includegraphics[width=1\linewidth]{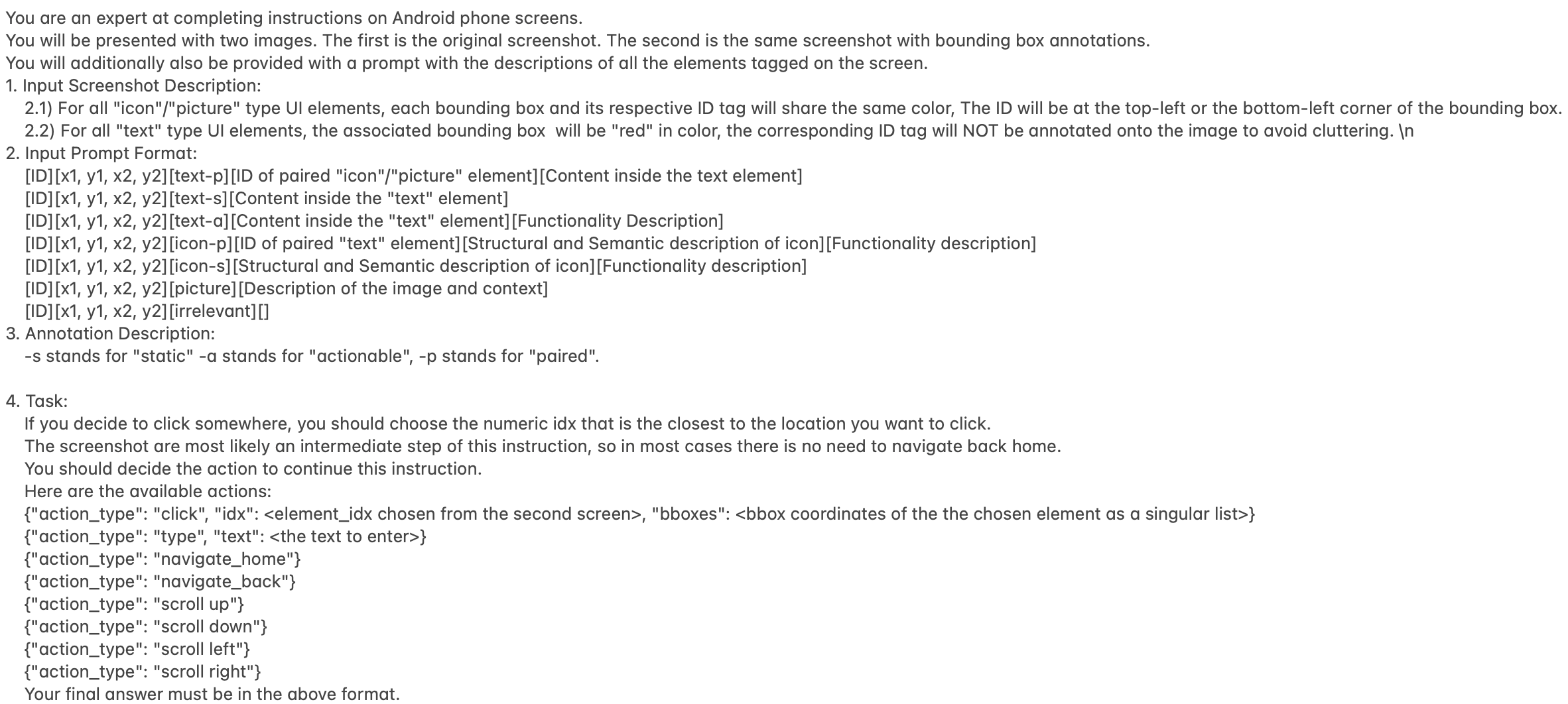}
    \caption{Agentic task following prompt for AITW}
    \label{fig:prompt_aitw}
\end{figure*}

\begin{figure*}
    \centering
    \includegraphics[width=1\linewidth]{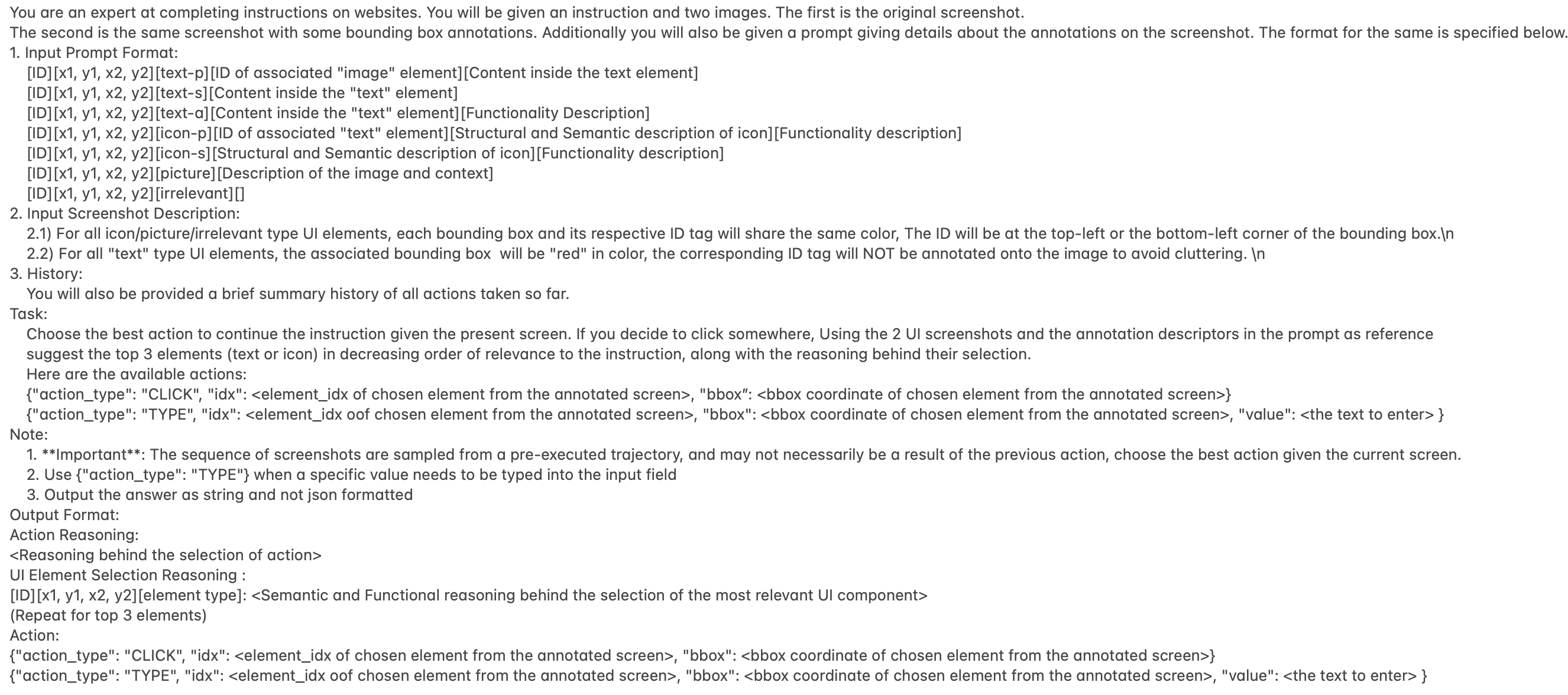}
    \caption{Agentic task following prompt for Mind2Web}
    \label{fig:prompt_mind2web}
\end{figure*}

\end{document}